%% file: main.tex
\documentclass[10pt]{article} % For LaTeX2e
\usepackage[preprint]{tmlr}
\usepackage{graphicx}
\input{math_commands.tex}

\usepackage{amsmath}
\usepackage{hyperref}
\usepackage{url}
\usepackage{enumitem}
\usepackage{booktabs}
\usepackage{fvextra}
\usepackage{enumitem}
\usepackage[most]{tcolorbox}
\usepackage[table]{xcolor}
\usepackage{pgfmath}   % or \usepackage{tikz}, which loads it

\newif\ifheatmap
\heatmaptrue                     % set to \heatmapfalse to turn coloring off
\definecolor{heatlow}{RGB}{220,60,60}
\definecolor{heathigh}{RGB}{60,170,80}

\newcommand{\hc}[2]{%
  \ifheatmap
    \pgfmathsetmacro{\heatval}{max(0.3,min(0.7,#1))}%
    \ifdim\heatval pt<0.5pt
      \pgfmathsetmacro{\heatpct}{120*(1-2*\heatval)}%
      \edef\heatcmd{\noexpand\cellcolor{heatlow!\heatpct!white}}%
    \else
      \pgfmathsetmacro{\heatpct}{120*(2*\heatval-1)}%
      \edef\heatcmd{\noexpand\cellcolor{heathigh!\heatpct!white}}%
    \fi
    \heatcmd
  \fi
  $#1 \pm #2$%
}
\newcommand{\hcin}[2]{\begin{tabular}[c]{@{}c@{}}\hc{#1}{#2}\end{tabular}}

\hypersetup{
  colorlinks=true,
  linkcolor=blue,
  citecolor=blue,
  urlcolor=blue,
}

\title{Pyrite: Building and Benchmarking \\ Protein-Ligand Pose Ranking Agents}

\author{\name Megha Srivastava\thanks{Work done during an internship at Genesis Molecular AI.}
  \email megha@cs.stanford.edu \\
  \addr Stanford University
  \AND
  \name Michael H. Lim
  \email mlim@genesis.ml \\
  \addr Genesis Molecular AI
  \AND
  \name Allison M. Keys
  \email allikeys@genesis.ml \\
  \addr Genesis Molecular AI
  \AND
  \name Pranav Murugan
  \email pranav@genesis.ml \\
  \addr Genesis Molecular AI
  \AND
  \name Nate Gruver
  \email ngruver@genesis.ml \\
  \addr Genesis Molecular AI
}

\def\month{MM}  % Insert correct month for camera-ready version
\def\year{YYYY} % Insert correct year for camera-ready version
\def\openreview{\url{https://openreview.net/forum?id=XXXX}} % Insert correct link to OpenReview for camera-ready version

\begin{document}

\maketitle

\begin{abstract}
Predicting how a ligand binds to its target is central to small-molecule drug discovery. Modern cofolding models can generate diverse candidate poses, but reliably ranking these candidates remains challenging. We introduce \textbf{Pyrite}, a pairwise pose-ranking benchmark comprising 1,500 pose pairs across diverse protein–ligand complexes. Expert chemists equipped with 3D visualization tools and physics-based scores perform near chance, highlighting the difficulty of interpreting available chemical signals. To investigate automated ranking, we introduce \textbf{PyriteRep}, a structured representation of molecular geometry, protein–ligand interactions, and physicochemical scores, and develop two LLM-based agents: \textbf{PyriteReasoner}, which directly compares poses through text-based reasoning, and \textbf{PyriteFeaturizer}, which generates executable code to construct features for logistic regression, optionally incorporating features derived from LLM subcalls. These agents achieve up to 67\% and 74\% accuracy, respectively, exceeding observed expert performance. LLM-guided feature discovery generally outperforms direct reasoning, while Group Relative Policy Optimization (GRPO) improves open-weight models, enabling them to match or surpass closed-source frontier models in our evaluations. These results establish \textbf{Pyrite} as a challenging test of pose-ranking ability and highlight LLM-guided feature discovery as a promising approach to interpret and combine the outputs of computational chemistry tools.
\end{abstract}

\begin{center}
\textbf{Code}: \url{https://github.com/genesistherapeutics/pyrite}\\
\textbf{Dataset}: \url{https://huggingface.co/datasets/genesisml/pyrite}
\end{center}

\input{introduction}

\input{setup}

\input{benchmark}
\input{humans}

\input{method}

\input{results}

\input{related}
\input{conclusion}

\medskip

\bibliography{neurips}
\bibliographystyle{tmlr}

\appendix
\section*{Appendix}

\input{appendix}

\end{document}

%% file: math_commands.tex
\usepackage{amsmath,amsfonts,bm}

\def\eqref#1{equation~\ref{#1}}
\def\1{\bm{1}}

\DeclareMathAlphabet{\mathsfit}{\encodingdefault}{\sfdefault}{m}{sl}
\SetMathAlphabet{\mathsfit}{bold}{\encodingdefault}{\sfdefault}{bx}{n}

%% file: introduction.tex
\section{Introduction}
Structure-based drug design typically begins from an experimental crystal structure, but because new structures are slow and costly to obtain, the field relies on computational pocket discovery, docking, and scoring \citep{Batool2019}. State-of-the-art cofolding models~\citep{Abramson2024, zhang2026protenix, pearl}, combined with diversity-promoting sampling, can now predict many plausible binding modes for a protein-ligand pair. However, this diversity is only useful if the most plausible poses can be identified. At test time, the true binding mode is unknown, and chemists must triage large sets of predictions. Errors in pose ranking can waste downstream effort and erode trust in the underlying models. We study whether agentic AI systems offer a potential path toward faster and more accurate pose ranking.

Pose ranking is difficult because the available signals are individually weak and often disagree. Physics-based scores such as MM-GBSA~\citep{genheden2015mmpbsa}, an estimate of binding free energy, and ligand torsional strain from Mogul~\citep{bruno2004mogul} each correlate only loosely with distance to the native pose, and for a given pair they may favor different candidates. Chemists resolve these conflicts using domain knowledge, such as familiarity with the target or chemical series, and by inspecting structures in 3D visualization software. For AI agents, this highlights a central challenge: invoking tools is easy, but interpreting their conflicting outputs to reach a sound decision is not. \textit{Can a language model-based agent, given the same computational tools as a chemist, rank poses more accurately than either the tools alone or expert chemists?}

In this work, we create and publicly release \textbf{Pyrite}, the first dedicated benchmark for pairwise pose ranking of protein--ligand complexes. The limited pose diversity of many publicly available cofolding models reduces the utility of physics-based scores for ranking and provides less challenging decoys for benchmarking (Figure~\ref{fig:placeholder}). This is consistent with recent work~\citep{khiari2026agenticposesrankeragenticaiframework}, which combines an LLM agent with physicochemical tools to rank poses from classical docking, but only matches, rather than exceeds, the underlying scoring function's accuracy on a small benchmark. In contrast, \textbf{Pyrite} draws poses from our proprietary cofolding model and diversity-promoting sampling algorithm, yielding a challenging benchmark of 1,500 pose pairs spanning 285 structures and 1,799 unique poses. 

We use \textbf{Pyrite} to investigate whether LLM-based agents can effectively interpret and combine the outputs of computational chemistry tools. Surprisingly, expert chemists perform near or below random chance despite substantial agreement between annotators, suggesting that conventional chemical intuition can be misleading when distinguishing high-quality cofolded poses. To systematically evaluate automated approaches, we next develop two complementary LLM-based agents: \textbf{PyriteReasoner} directly compares candidate poses through explicit text-based chemical reasoning, while \textbf{PyriteFeaturizer} uses an LLM to discover informative structural features for a lightweight logistic-regression classifier. This distinction allows us to investigate whether LLMs are more effective as direct pose-ranking agents or as feature engineers for conventional statistical models. Finally we explore whether reinforcement learning (RL) can improve the pose-ranking capabilities of open-weight models. Our contributions include:

\begin{enumerate}
\small
    \item \textbf{Pyrite}, a challenging pairwise pose-ranking benchmark constructed from diverse, high-quality cofolded poses. We find that expert chemists perform near or below random chance, despite substantial inter-annotator agreement, and that conventional physics-based scores can be anti-correlated with correctness.

    \item \textbf{PyriteRep}, a structured data format for representing 3D poses, including  molecular geometry, protein--ligand interactions, and precomputed physicochemical features. We demonstrate that \textbf{PyriteRep} substantially improves LLM-based agents' pose-ranking accuracy compared to raw 3D coordinates alone.

    \item \textbf{PyriteReasoner}, an LLM-based agent that directly ranks pose pairs through text-based chemical reasoning, achieving up to \textbf{67\%} accuracy. RL improves the performance of open-weight models, narrowing the gap with closed-source frontier models.

    \item \textbf{PyriteFeaturizer}, an LLM-based coding agent that discovers and implements predictive structural features for a simple logistic-regression classifier. By combining code-derived quantitative descriptors with features generated through LLM subcalls, \textbf{PyriteFeaturizer} achieves up to \textbf{74\%} pose-ranking accuracy, the highest performance on our benchmark. RL substantially improves feature discovery in open-weight models, enabling them to match or exceed closed-source frontier models.
\end{enumerate}

Overall, our results demonstrate that ranking poses generated by modern cofolding models remains a challenging task, even for experienced chemists. While classical models trained on quantitative structural features are surprisingly competitive with direct LLM reasoning, LLM-guided feature discovery offers a promising way to improve upon these models by incorporating additional chemical and biological knowledge.

\begin{figure}[t]
    \centering
    \fbox{\includegraphics[width=0.85\linewidth]{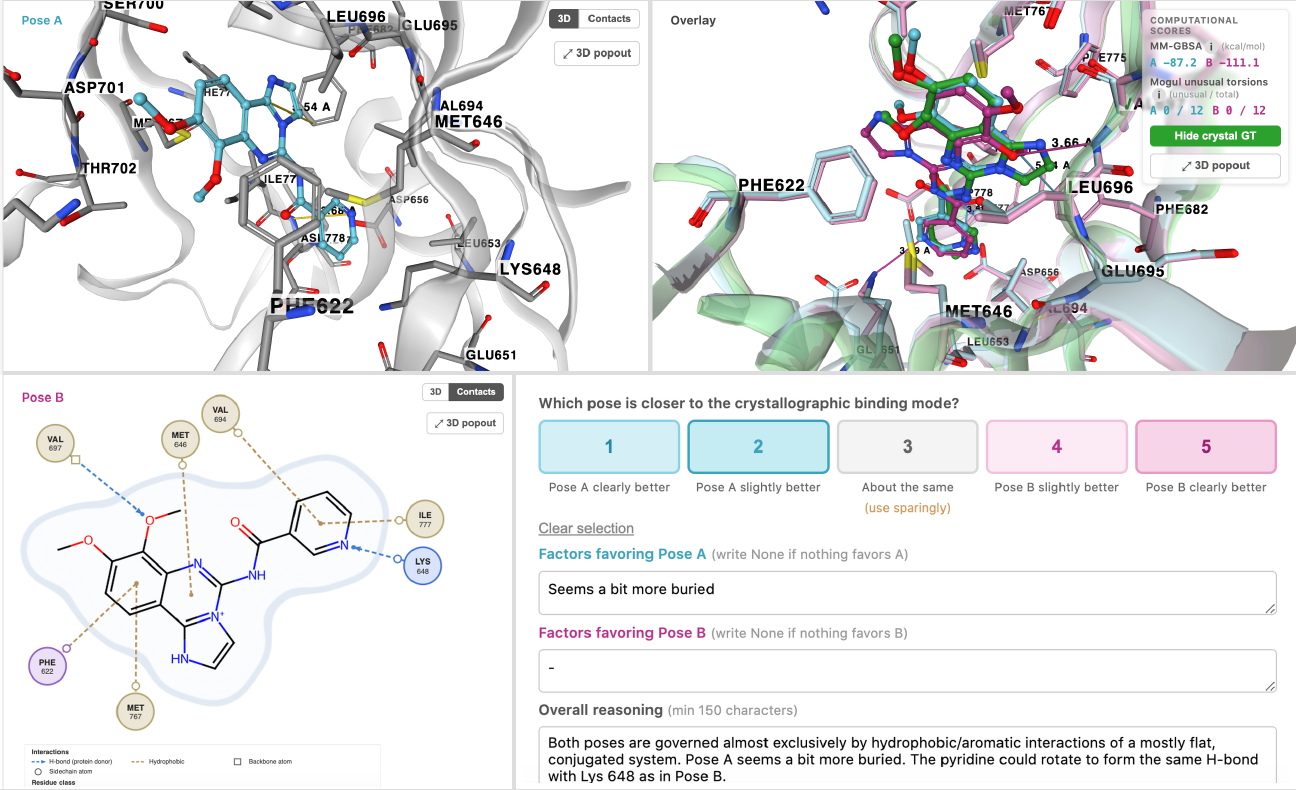}}
    \caption{Interface for human expert annotations on our pairwise pose ranking benchmark \textbf{Pyrite} (ground truth crystal shown only for paper visualization). The example shows two predicted poses of a potential small-molecule inhibitor of the autotaxin protein (\href{https://www.rcsb.org/structure/7G63}{Protein Data Bank 7G63}), with toggle control  for 2D contact map and interactive 3D and overlay views. Experts are asked to select which pose is closer to the crystallographic binding mode (selecting 1-5), write factors in favor of both pose A and  B, and describe their reasoning on how they weigh these factors for pose selection. Annotators are also provided MM-GBSA scores and the \# of unusual torsions using Mogul for each pose.}
    \label{fig:interface}
\end{figure}

%% file: setup.tex
\section{Problem Setup: Pose Ranking}
\label{feature}
To introduce the pose ranking problem, let us walk through the example in Figure \ref{fig:interface}, which shows two predicted poses (Poses A and B) of a small-molecule that binds to autotaxin protein (\href{https://www.rcsb.org/structure/7G63}{Protein Data Bank 7G63}). These predictions come from a proprietary cofolding model. At test time, the ground truth crystallographic binding mode of the protein-ligand complex is unknown; therefore, a strong pose ranker should be able to select which pose is closer to the ground truth 3D structure.  In the example shown, the correct choice is Pose A. Now, consider the following reasoning from an expert chemist: \begin{quote}
    \textbf{Expert Reasoning: }\textit{Both poses are governed almost exclusively by hydrophobic/aromatic interactions of a mostly flat, conjugated system. Pose A seems a bit more buried. The pyridine could rotate to form the same H-bond with Lys 648 as in Pose B.}
\end{quote}

The chemist weighs many different factors before ultimately choosing Pose A, combining physical reasoning with more intuitive features. Our goal is to  \textit{learn} a strong pose ranker, and investigate whether AI agents based on language models can successfully reason over the outputs of computational chemistry tools to achieve high pose ranking  accuracy. Although strong pairwise pose ranking ability is not the same as overall ranking, both tasks share the same underlying reasoning requirements from AI agents. We specifically chose to focus on a pairwise pose ranking due to its simplicity in light of context-length limits of language models, as well as the cognitive burden placed on the human experts we use for evaluating our benchmark. Furthermore, pairwise comparison allows us to derive better semantic features than scoring a single pose.

\subsection{Representing poses for AI agents with PyriteRep}
\label{features-app}
Inspired by recent work showing that molecular graph schemas help improve the performance of LLM-based systems for chemical discovery \citep{runcie2026molecularrepresentationslargelanguage},  we introduce \textbf{PyriteRep}, a structured data format for representing 3D poses for our pose ranking task.  In Section \ref{sec:classical}, we will show that \textbf{PyriteRep}  improves the performance of LLM-based agents reasoning in text over pose ranking, in comparison to only providing each pose's 3D-coordinates.   \textbf{PyriteRep} consists of 29 features, including both \emph{pose-invariant} (identical for both poses) and \emph{pose-dependent} features. These features are the outputs of several computational chemistry tools, including RDKit and OpenMM \citep{rdkit_2026_09_1b1,eastman2024openmm8}. We describe the features in the boxes below. Pose geometry features are computed from the ligand's $N$ heavy-atom 3D coordinates, with centroid $\bar{\mathbf{r}} = \frac{1}{N}\sum_i \mathbf{r}_i$ and gyration tensor $S = \frac{1}{N}\sum_i (\mathbf{r}_i - \bar{\mathbf{r}})(\mathbf{r}_i - \bar{\mathbf{r}})^\top$.

\begin{tcolorbox}[
    colback=violet!10,
    colframe=violet!50,
    fontupper=\small,
    title=Pose Invariant Features in PyriteRep,
    colbacktitle=violet!90,
    coltitle=white
]

\begin{enumerate}[leftmargin=5pt]
    \item \textbf{Target context}: Protein ID, name, family name, and description of function from \href{https://www.uniprot.org/}{UniProt}.
    \item \textbf{Atom List}: List of heavy atoms with element and index.
    \item \textbf{Aromatic NH atoms}: List of indices of hydrogen-bearing aromatic nitrogen atoms.
    \item \textbf{Bond List}: List of bonds with type [single (\texttt{-}), double (\texttt{=}), triple (\texttt{\#}), or aromatic (\texttt{:})].
    \item \textbf{Formal Charges}: List of non-zero formal charges per atom.
    \item \textbf{SMILES}: Canonical SMILES string \citep{weininger1989smiles}. 
    \item \textbf{IUPAC name}: Canonical IUPAC name \citep{favre2014nomenclature}.
    
\end{enumerate}

\end{tcolorbox}

\begin{tcolorbox}[
    colback=violet!10,
    colframe=violet!50,
    fontupper=\small,
    title=Pose Dependent Features in PyriteRep,
    colbacktitle=violet!90,
    coltitle=white
]

\begin{enumerate}[leftmargin=5pt]
\item \textbf{3D Coordinates}: List of 3D coordinates by (id, x, y, z).
\item \textbf{Asphericity}: $b = \lambda_3 - \frac{1}{2}(\lambda_1 + \lambda_2)$, from eigenvalues $\lambda_1 \le \lambda_2 \le \lambda_3$ of $S$; $b = 0$ for a sphere.
    \item \textbf{Radius of Gyration}: $R_g = \sqrt{\frac{1}{N}\sum_i \lVert \mathbf{r}_i - \bar{\mathbf{r}} \rVert^2} = \sqrt{\lambda_1 + \lambda_2 + \lambda_3}$.
    \item \textbf{Maximum Extent}: $D_{\max} = \max_{i,j} \lVert \mathbf{r}_i - \mathbf{r}_j \rVert$.
    \item \textbf{Interactions List}: List of interactions with residue, backbone/side-chain origin, and type (Hydrogen-bond Donor, Hydrogen-bond Acceptor, Salt Bridge, $\pi$-stacking, etc.).
    \item \textbf{\# of Internal Close Contacts}: $\sum_{i<j,\,|i-j|\ge 2} \mathbb{1}[\lVert \mathbf{r}_i - \mathbf{r}_j \rVert < d_{\text{cc}}]$,  $d_{\text{cc}} = 3.2$\,\AA\ .
    \item \textbf{\# Residues Contacted}: Total number of distinct residues making at least one detected contact.
    \item \textbf{\# Sidechain Residues Contacted}: Total number of distinct sidechain residues making at least one detected contact. 
    \item \textbf{Torsional Strain}: Torsion strain for the (i) protein backbone, (ii) sidechain, and (iii) total (kcal/mol).
    \item \textbf{Per-Residue Strain}: Torsion strain for the (i) protein backbone, (ii) sidechain, and (iii) total divided by $N_{\text{res}}$ (kcal/mol).
    \item \textbf{PoseBusters Validity}: \textsc{PoseBusters} validity check \citep{buttenschoen2024posebusters}.
    \item \textbf{Mogul Total \# Torsions}: Total \# of ligand torsion angles analyzed based on the Cambridge Structural Database (CSD) library using Mogul \citep{bruno2004retrieval}.
       \item \textbf{Mogul \% Unusual Torsions}:  \% of ligand torsion angles that are unusual, deviating from the CSD distribution, based using Mogul.
    \item \textbf{Mogul \# Unusual Torsions}: Total \# of ligand torsion angles that are unusual based using Mogul.
    \item \textbf{Mogul Torsion Penalty}: Aggregate deviation from CSD torsion angle distributions using Mogul.
    \item \textbf{Mogul \# Unusual Rings}: Total \# of ligand rings with conformations deviating from CSD geometries using Mogul.
    \item \textbf{Inter-pose ligand RMSD}: $\sqrt{\frac{1}{N}\sum_i \lVert \mathbf{r}^A_i - \mathbf{r}^B_i \rVert^2}$.
     \item \textbf{MM-GBSA Score}: Binding energy score, estimated from first step of docking pipeline.
    
\end{enumerate}
\end{tcolorbox}

%% file: benchmark.tex
\section{Pyrite Benchmark Creation}

\begin{figure*}
    \centering
    \includegraphics[width=0.99\linewidth]{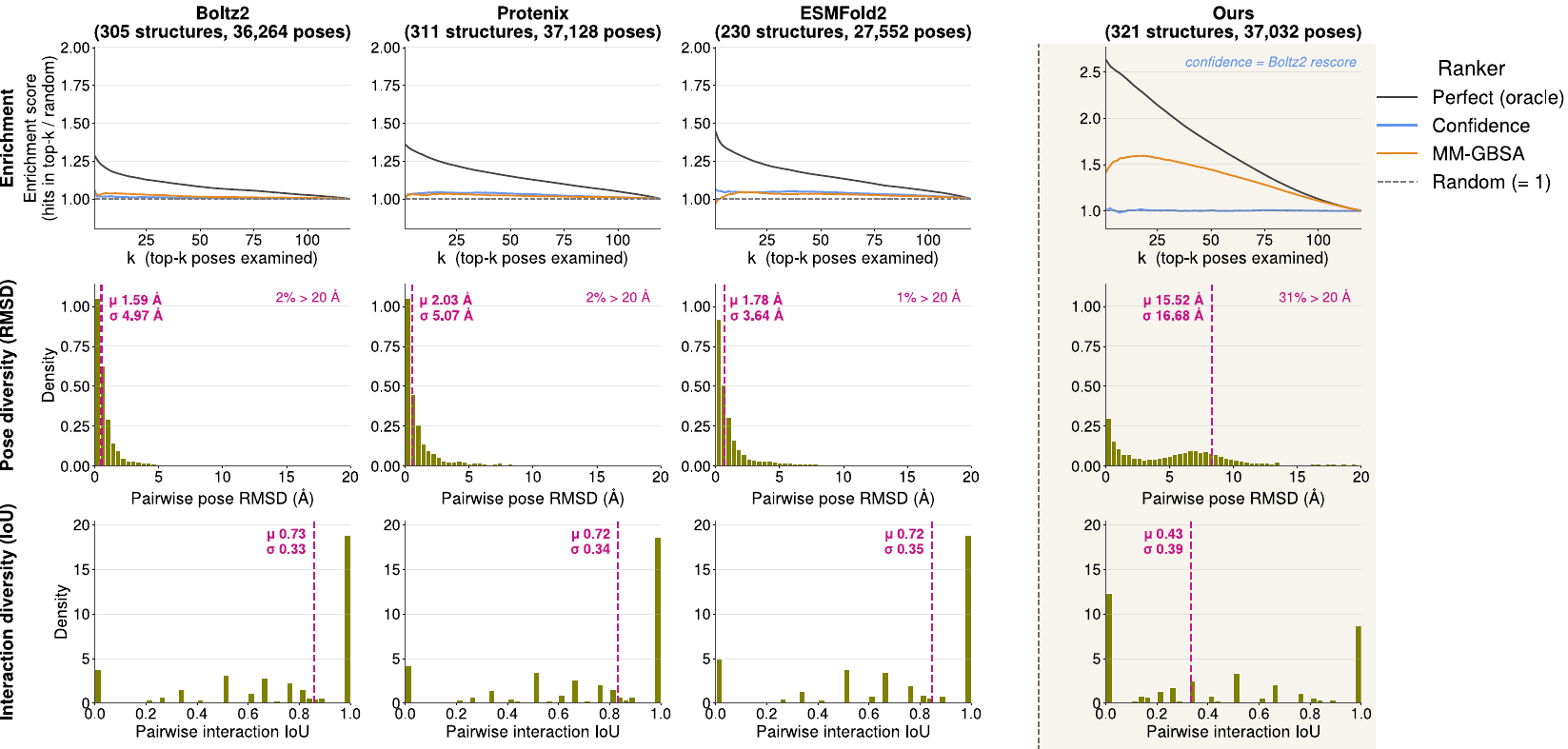}
    \caption{\textbf{Top-k pose enrichment and diversity across cofolding models} For each of Boltz2, Protenix, ESMFold2, and our cofolding model, we scored 120 poses per complex against the ground truth crystal, calling a pose a "hit" at RMSD < 2 \AA\;and lDDT-PLI $\geq$ 0.8. \textbf{Top Row} shows top-k enrichment (EF@k = hits in the top-k
over hits expected at random, pooled across complexes, so random = 1). \textbf{Middle Row} and \textbf{Bottom Row}: the 3rd-party cofolding models
are far less diverse than ours, measured both by the distribution of per-pair RMSD and Interaction IoU.}
    \label{fig:placeholder}
\end{figure*}

\subsection{Generating Diverse Poses}
We create \textbf{Pyrite}  using poses from our proprietary cofolding model and inference-time method that generates a diverse set of binding poses for each target, while using physics-based guidance to steer predictions toward chemically and geometrically valid structures. Figure \ref{fig:placeholder} shows that the poses from our model are more diverse than three different public 3rd-party models (Boltz2 \citep{passaro2025boltz2}, Protenix \citep{zhang2026protenix}, and ESMFold2 \citep{candido2026esmfold2}). Moreover, for all three cofolding models, both the model’s own confidence and MM-GBSA score barely passes random chance at pose ranking, falling significantly short of the oracle.  Only our system receives real enrichment from MM-GBSA, but that
underperforms a far higher ceiling, necessitating building stronger pose-ranking methods.

To create our benchmark, we first use our cofolding model to sample up to 120 poses for protein-ligand queries spanning more than 50,000 structures from the \cite{rcsb2026}.  We then take the top-25 poses scored by MM-GBSA for each query, entering a tight regime of high-quality poses where, as we will later show in  Figure \ref{fig:placeholder}, MM-GBSA can actually become anti-correlated with pose ranking performance. We then consider all possible pairs of poses for the same query, and filter this set to create \textbf{Pyrite}.

\subsection{Filtering Pose Pairs for Comparison}
\label{compare}
Not all pairs are useful for a pose ranking benchmark: for example, some poses may be obviously low quality or structurally invalid, while some pairs of poses might be too similar to each other to distinguish. We first categorize each pose by measuring  RMSD-to-ground truth crystal over the 3D coordinates for each atom,  lDDT-PLI (Local Distance Difference Test for Protein–Ligand Interactions), which is a scoring metric that measures local interatomic distances  \citep{robin2023assessment}, and the interaction recall given the reference crystal's interactions. We designate thresholds for these metrics in order to create three categories: 

\begin{tcolorbox}[
    colback=purple!10,
    colframe=purple!50,
    title=Pose Categorization (Ground Truth Crystal Reference),
    colbacktitle=purple!90,
    coltitle=white
]
\small

\begin{enumerate}[leftmargin=5pt]
\item \textbf{GOOD}: RMSD-to-ground truth crystal < 1 \AA, lDDT-PLI  $\geq 0.85$, interaction recall $\geq 0.90$
    \item \textbf{MAYBE}: RMSD-to-ground truth crystal < 2 \AA\;and $\geq$ 1 \AA, $ 0.5\leq$ lDDT-PLI  $< 0.85$ 
    \item \textbf{BAD}: RMSD-to-ground truth crystal < 4 \AA\;and $\geq 2$ \AA, lDDT-PLI $< 0.5$, interaction recall $< 0.5$ 
\end{enumerate}
\end{tcolorbox}

 We  filter down our dataset to only consist of \textbf{Hard} pairs (GOOD, MAYBE) and \textbf{Easy} pairs (GOOD, BAD), ensuring that every pair in \textbf{Pyrite} consists of at least one high-quality pose. However, we found that this high-level categorization was not sufficient in removing pairs with poses that were too similar or low-quality, so we introduced additional filtering conditions described in Appendix \ref{sec:filter}. After applying all these filters, our final \textbf{Pyrite} benchmark consists of 1,500 pose pairs spanning 285 unique structures and 1,799 unique poses. Of these structures, 195 were released in the RCSB Protein Data Bank before September 30, 2021 (\textit{pre-AF3}) and 90 were released afterwards (\textit{post-AF3}). Because human annotation is costly, we select a \textit{curated} subset of 750 pairs (177 structures, 890 poses), on which we collect annotations from expert chemists (see Section \ref{sec:humans}). For the rest of this paper, we will focus on three splits within \textbf{Pyrite}:
\begin{itemize}
    \item \textit{pre-AF3 curated}: 485 pairs, 121 structures. We report human accuracy on this split in Section \ref{sec:humans}. We do not evaluate LLM agents (see Sections \ref{sec:reason} and \ref{sec:code}) on pre-AF3 pairs, since we reserve the pre-AF3 data for training data
    \item \textit{post-AF3 all}: 484 pairs, 90 structures, every post-AF3 pair in \textbf{Pyrite}. 
    \item \textit{post-AF3 curated}: 265 pairs, 56 structures, the subset of \textit{post-AF3 all} with human annotations. This is the only split where we can directly compare LLM agents against expert chemists.
\end{itemize}

Because \textbf{Pyrite} is a balanced binary-classification task, we focus on comparing the \textit{accuracy} of various pose-ranking methods. To ensure fair comparison, we treat abstentions (neither choosing Pose A or B) as incorrect across all methods.

\subsection{Physics scores are anti-correlated for strong pose decoys}
We first check whether the pre-computed physics scores we include in \textbf{PyriteRep} (see Section \ref{features-app}) can, as individual features, serve a strong pose rankers on \textbf{Pyrite}. For example, the MM-GBSA score estimates the binding free energy of a protein-ligand complex by combining molecular mechanics energies with implicit solvation models; intuitively, a lower MM-GBSA structure can imply that the predicted pose is more thermodynamically stable and closer to the native binding mode. However, we find that \textit{selecting the pose with the lower MM-GBSA score is anti-correlated with the correct choice} in \textbf{Pyrite} (see Table \ref{tab:classic}), leading to low accuracy. This is likely due to \textbf{Pyrite} consisting of high-quality pose decoys that are predicted to be thermodynamically stable by construction, and therefore MM-GBSA is not a reliable score for pose ranking in this regime. 
In fact, Figure \ref{fig:placeholder} shows no ranking improvement when k < 25, and a slight dip in the k < 10 regime of enrichment scores when using MM-GBSA for ranking poses from our system.  

Importantly, no aspect of our benchmark creation intentionally selected for pose pairs where MM-GBSA provided an incorrect signal; rather, this appears to be a property of the high-quality pose regime for cofolding. Table \ref{tab:classic} shows that other scores, including the total $\#$ of interactions, total torsion strain, and total $\#$ of unusual torsion angles using Mogul, similarly result in poor pose ranking accuracy across all splits. Collectively, these results suggest that any strong pose ranking method needs to both (i) overcome any potential inductive biases (e.g. that lower MM-GBSA scores are more favorable), and (ii) learn non-trivial ways to combine all the features in \textbf{PyriteRep}.

Given these results, a natural question is whether one should simply invert the sign of a physical score like MM-GBSA to use as the ranking criterion, which would by construction achieve above-chance accuracy on \textbf{Pyrite}. This would be problematic as such a rule contradicts the physical interpretation of the score, and would be inadvisable to deploy without evidence that the inversion generalizes beyond our benchmark: in fact, Figure \ref{fig:placeholder} shows that the relationship between MM-GBSA and pose correctness depends on the number of candidate poses retained. Nevertheless, we find empirically that our trained methods presented later in Section \ref{sec:results} outperform sign-flipped MM-GBSA with a low agreement-rate (Appendix \ref{sec:app-flip}), indicating that it is possible to learn a stronger decision rule that is not just a noisy copy of inverted physical heuristics.

\begin{table}[t]
    \centering
    \caption{Accuracy  and 95\% CIs from expert chemists, Boltz-2 confidence, and various physics scores from \textbf{PyriteRep} across three benchmark splits. No method achieves pairwise accuracy significantly above chance. 
    }
    \vspace{5pt}
    \label{tab:classic}
    \small
\begin{tabular}{lccc}
        \toprule
        Predictor & Pre-AF3 curated & Post-AF3 curated & Post-AF3 full \\
        \midrule
        Human experts (abstain-as-wrong)   & \hc{0.42}{0.03} & \hc{0.45}{0.04} &- \\
        Human experts (decisive only)      & \hc{0.51}{0.03} & \hc{0.57}{0.05} & - \\
        MM-GBSA energy $\downarrow$        & \hc{0.32}{0.04} & \hc{0.34}{0.06} & \hc{0.35}{0.04} \\
        Total interactions $\uparrow$      & \hc{0.41}{0.04} & \hc{0.40}{0.06} & \hc{0.40}{0.04} \\
        Total torsion strain $\downarrow$  & \hc{0.45}{0.04} & \hc{0.51}{0.06} & \hc{0.51}{0.04} \\
        Mogul unusual torsions $\downarrow$ & \hc{0.35}{0.04} & \hc{0.50}{0.06} & \hc{0.43}{0.04} \\
        Boltz-2 confidence $\uparrow$      & \hc{0.52}{0.04} & \hc{0.46}{0.06} & \hc{0.45}{0.04} \\
        \bottomrule
    \end{tabular}
\end{table}

%% file: humans.tex
\section{Human expert performance: oracle or baseline?}
\label{sec:humans}

We next evaluate \textbf{Pyrite} with expert human chemists. Although our initial goal was to use the reasoning text provided by human experts to guide the design a strong automated pose-ranking method, we surprisingly find that expert chemists perform below/at random chance on \textbf{Pyrite}, as we describe below. 

\subsection{Human expert annotation overview}

We recruit 5 expert chemists with an average of more than 10 years of experience with computer-aided drug discovery (CADD), molecular docking, and virtual screening. Each annotator was provided a set of pose pairs, and asked to choose which pose (A or B) is closer to the crystallographic binding mode for each pair, providing the following: 
\begin{enumerate}
    \item \textbf{Selection}: A choice from 1-5 corresponding to which pose the expert believes is closer to the crystallographic binding mode (1: Pose A clearly better, 2: Pose A slightly better, 3: About the same, 4: Pose B slightly better, 5: Pose B clearly better).  
    \item \textbf{Factors favoring Pose A/B:} Any specific factors (or "None") about both Pose A and B that make it closer to the crystallographic binding mode. 
    \item \textbf{Overall Reasoning}: The overall reasoning behind the selection,  describing all features of the two poses that influenced your decision, as well as any trade-offs. For the reasoning, we enforce a minimum 150 character requirement.
\end{enumerate}

Figure \ref{fig:interface} shows the interface provided to human experts, as well as example annotations.  Experts were provided both 3D visualizations and 2D contact maps, as well as two physics-based and statistical scores: MM-GBSA and Mogul \# of Unusual Torsions. Furthermore, experts were provided the UniProt protein family name and description, and could download PDB/SDF files of the protein-ligand poses to further analyze in molecular visualization tools like PyMOL~\citep{PyMOL}. We collected 2 annotations per pose pair in the \textit{curated} subset of \textbf{Pyrite}, reaching a total of 1500 annotations.

\subsection{Human experts perform below chance at pairwise ranking of cofolded poses}
 We surprisingly find that human experts achieved an average pose pair comparison accuracy below chance for both pre-AF3 and post-AF3 structures (see Table \ref{tab:classic}). This low performance holds even when we removed pose pairs where annotators selected option 3 (\textit{``About the same''}, \textit{pre-AF3} $0.42 \rightarrow 0.51$, \textit{post-AF3} $0.45 \rightarrow 0.57$). Interestingly, the agreement rate between annotators (\textbf{77\%})  was significantly higher than their accuracy. 

\begin{figure}
    \centering
    \includegraphics[width=0.99\linewidth]{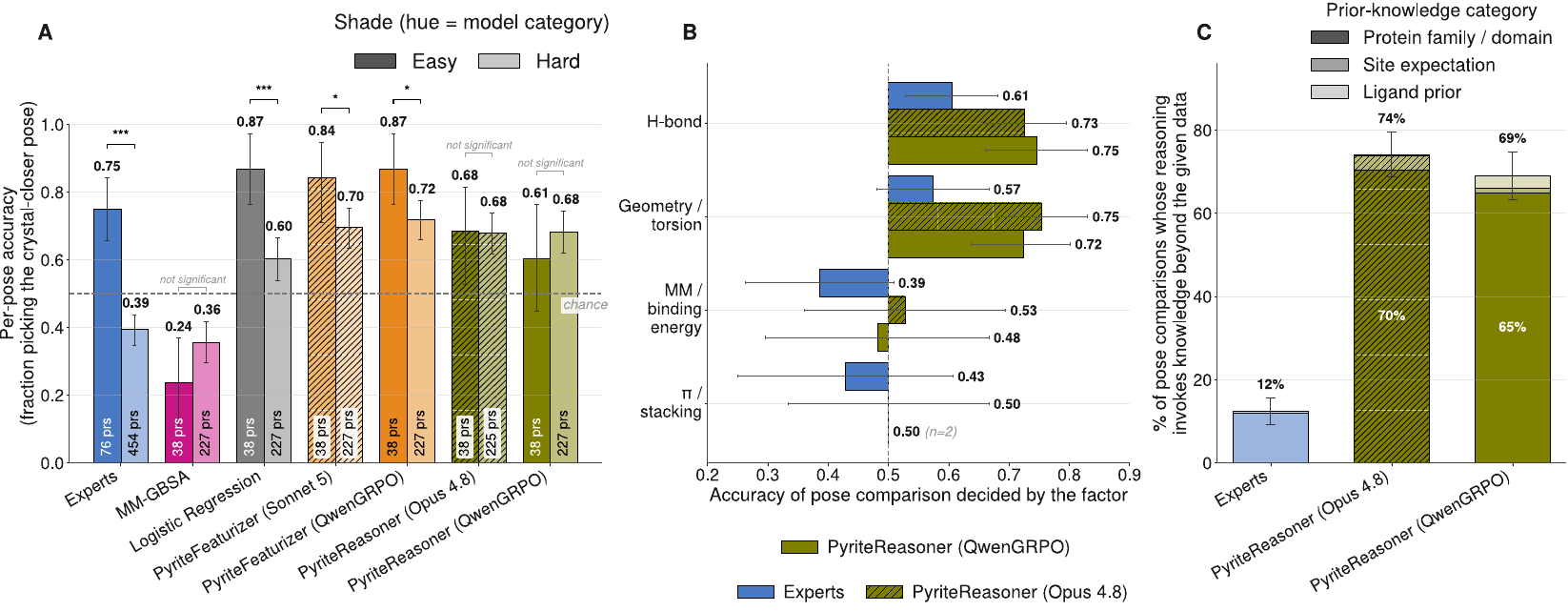}
    \caption{\textbf{Human experts vs.\ models performance on post-AF3 curated split. (A)} Accuracy for experts and a subset of ranking methods, split by \textbf{Easy} and \textbf{Hard} pairs. Error bars are  95\% CIs. \textbf{(B)} Average accuracy across experts and two \textbf{PyriteReasoner} agents grouped by factors driving the reasoning for a given pair selection. \textbf{(C)} Rate at which experts and and two \textbf{PyriteReasoner} agents reference protein-family and domain knowledge in their reasoning output.}
\label{fig:human}
\end{figure}

One hypothesis for human expert's poor accuracy on our benchmark is that pairs based on poses generated from cofolding models are, versus classical docking methods, more likely to have subtle differences that are unintuitive for experts to detect, despite their extensive molecular docking experience. In fact, we observe a statistically significant gap in expert accuracy between the \textbf{Easy} and \textbf{Hard} pairs in our benchmark (see Figure \ref{fig:human}), suggesting that experts achieve higher accuracy when the difference between two poses is more significant. We also find that experts often rely on the MM-GBSA score for their selection which, as shown earlier, is actually anti-correlated with accuracy on \textbf{Pyrite}. In fact, expert accuracy is low when their reasoning relies on the MM-GBSA score, but is significantly higher whenever they mention features like the pose's interactions or torsion geometry (Figure \ref{fig:human}). Overall, the surprisingly poor performance of experts chemists on \textbf{Pyrite} motivates the need for building stronger automated pose ranking methods.

% 4000 bootstrap samples, clustered by structure. 0.001, 0.01, 0.05 as star values

%% file: method.tex
\section{Methods: Building pose ranking agents}

In this section, we give an overview of our two LLM-based agents for pose ranking: \textbf{PyriteReasoner} and \textbf{PyriteFeaturizer}. Recent advances in large language models (LLMs) have enabled the development of agents capable of performing complex scientific tasks, including chemical synthesis planning, drug discovery, and autonomous experiment design \citep{boiko2023autonomous,bran2024augmenting,roohani2025biodiscoveryagent}. These advances motivate our exploration of LLM-based agents for pose ranking, a task that requires integrating qualitative chemical reasoning with quantitative assessments of molecular geometry, interactions, and physics.

A key motivation for our approach is the potential to combine the interpretability of language-based reasoning with the precision of quantitative analysis. While LLMs are increasingly capable of articulating chemical principles and providing human-readable explanations for their predictions, their reasoning may also inherit the heuristics and biases of human expert chemists (e.g., the assumption that lower MM-GBSA scores necessarily indicate better poses). Meanwhile, exposure to high-quality poses from \textit{pre-AF3} structures may help counter such biases by grounding agents' reasoning in empirical structural evidence, enabling them to identify useful pose-ranking criteria that extend beyond conventional chemical intuition.

We therefore develop two complementary approaches. \textbf{PyriteReasoner} uses an LLM to directly evaluate candidate poses through explicit chemical reasoning. In contrast, \textbf{PyriteFeaturizer} draws inspiration from automated feature engineering \citep{katz2016explorekit, kanter2015deep}, using an LLM to discover informative structural features that can be incorporated into a lightweight logistic-regression classifier. Together, these approaches allow us to investigate the relative merits of using LLMs as direct reasoning agents versus using them to discover features for conventional models.

\subsection{Text-based reasoning with PyriteReasoner}

\textbf{PyriteReasoner} compares candidate pose pairs through explicit text-based reasoning, leveraging chemical and structural knowledge acquired during LLM pretraining to assess their relative plausibility. To examine how model capabilities affect pose-ranking performance, we instantiate \textbf{PyriteReasoner} with four LLMs: two large frontier models (Opus 4.8 and GPT-5.6-sol), one midsize frontier model (Sonnet 5), and one open-weight model (Qwen3.8-27B). We evaluate five input ablations to investigate how different representations of structural information affect pose-ranking performance:
\begin{enumerate}
\small
    \item \textbf{Raw 3D Coordinates Only}: Protein and ligand atomic coordinates for both poses.
    \item \textbf{Interactions + Raw 3D Coordinates}: Atomic coordinates supplemented with a list of protein--ligand interactions for each pose.
    \item \textbf{PyriteRep w/ 3D Coordinates}: All features in \textbf{PyriteRep} (Section \ref{feature}), including atomic coordinates.
    \item \textbf{PyriteRep w/o 3D Coordinates}: All \textbf{PyriteRep} features except atomic coordinates.
    \item \textbf{In-Context}: \textbf{PyriteRep} supplemented with two labeled examples retrieved from the \textit{pre-AF3} split, selected by ligand Tanimoto similarity to the query.
\end{enumerate}

For ablations using \textbf{PyriteRep}, the large number of features may make it difficult for the LLM to identify the most informative differences between poses. To make these differences more salient, we append a block of precomputed summaries derived from \textbf{PyriteRep}, explicitly highlighting features that differ between Pose A and Pose B, such as the shared interactions or overall number of hydrogen-bond donors (Appendix \ref{sec:precompute}). An example prompt for the \textbf{PyriteRep} w/o 3D Coordinates configuration is provided in Appendix \ref{prompt-reasoner}.

\subsection{Discovering predictive features with PyriteFeaturizer}
\label{method:regression}

Since many features in \textbf{PyriteRep} are quantitative, we hypothesize that classical machine learning methods, such as logistic regression, can learn which structural features are predictive of pose correctness, including relationships that may contradict the heuristics we saw experts employ (e.g. overreliance on MM-GBSA). However, LLMs may possess additional chemical and biological knowledge that is not readily captured by predefined quantitative descriptors, such as known interactions within protein families, or chemical intuition about which ligand substructures should be solvent-exposed or occupy hydrophobic pockets.

Motivated by prior work on automated feature engineering \citep{katz2016explorekit,kanter2015deep} and AutoML \citep{real2020automl}, we develop \textbf{PyriteFeaturizer}, an LLM-based coding agent that generates Python scripts to construct predictive features from the \textbf{PyriteRep} schema. These features are subsequently used to train a logistic-regression classifier. All classifiers are trained exclusively on structures from the \textit{pre-AF3} split. We evaluate \textbf{PyriteFeaturizer} using the same four LLMs as \textbf{PyriteReasoner} (Opus 4.8, GPT-5.6-sol, Sonnet 5, and Qwen3.8-27B) under three feature-generation ablations:

\begin{enumerate}
\small
    \item \textbf{Code Only}: \textbf{PyriteFeaturizer} generates Python code that computes features directly from \textbf{PyriteRep}, without additional LLM calls.
    \item \textbf{Protein LLM Subcall}: In addition to code-derived features, \textbf{PyriteFeaturizer} can design and invoke calls to a smaller LLM (Qwen3.5-9B) to extract protein-level features shared by Pose A and Pose B.
    \item \textbf{Pose LLM Subcall}: \textbf{PyriteFeaturizer} can additionally design and invoke calls to Qwen3.5-9B to extract pose-specific features for Pose A and Pose B.
\end{enumerate}

We compare these ablations against a \textbf{Vanilla Logistic Regression} baseline trained on all quantitative features in \textbf{PyriteRep}. We use logistic regression as our downstream classifier for its simplicity, but also evaluate alternative classical methods, including tree-based models, in Appendix \ref{sec:classical}.

\subsection{Improving pose ranking accuracy with GRPO} \label{method:grpo}
For both  \textbf{PyriteReasoner} and \textbf{PyriteFeaturizer}.  we investigate whether GRPO \citep{shao2024deepseekmath}, the de facto standard algorithm for RL-based reasoning improvements in open-weight models \citep{guo2025deepseekr1}, can similarly improve pose ranking accuracy on the \textit{post-AF3} evaluation splits. For both agents, we use the same \textit{pre-AF3} training dataset used in \ref{method:regression}, as well as the same base model (Qwen3.8-27B).  The rewards we use for training both agents are:  

\begin{itemize}
\small
    \item \textbf{PyriteReasoner}: An output (pose selection and reasoning) receives 0 unless it ends with a parseable \texttt{SELECTION} line, 1 if it is well-formatted but selects the incorrect pose, and 2 if it selects the correct pose. 
    \item \textbf{PyriteFeaturizer}: An output (a Python featurizer script) receives 0 if it fails to execute, exceeds 20s, or yields only constant features. Otherwise, the reward is the accuracy of a logistic regression trained on the returned features, under a 5-fold leave-structure-out cross-validation over the \textit{pre-AF3} training pool. 
\end{itemize}

Because training \textbf{PyriteFeaturizer} with GRPO optimizes the quality of a single Python featurizer script, controlling batch variation is a bigger challenge for \textbf{PyriteReasoner}, due to both sampling and input diversity.  More details on the training hyperparameters and implementation are in Appendix \ref{grpo-hparam}.

%% file: results.tex
\begin{table}[t]
    \centering
\caption{\textbf{PyriteReasoner} pose ranking accuracy across four different base LLMs, five input ablations, and  Qwen3.8-27B finetuned with GRPO. Values are mean accuracy $\pm$ 95\% confidence intervals. $\dagger$ means significantly higher than expert, curated split only, and * means significant difference for $\Delta$ GRPO, $p < 0.05$.} 
\vspace{5pt}
    \label{tab:llm_zeroshot}
    \small
    \setlength{\tabcolsep}{4pt}
    \begin{tabular}{lcccccc}
        \toprule
        Variant & Opus 4.8 & Sonnet 5 & GPT-5.6-sol & Qwen3.8-27B & Qwen3.8-27B-GRPO & $\Delta$ GRPO \\
        \midrule
        \multicolumn{7}{l}{\textit{Curated} ($n$=265)} \\
        \addlinespace[2pt]
        Raw 3D Coord Only            & \hc{0.49}{0.06} & \hc{0.34}{0.07} & \hc{0.52}{0.06}$^{\dagger}$ & \hc{0.53}{0.06} & \hc{0.52}{0.06}$^{\dagger}$ & $-0.01$ \\
        Interactions + Raw 3D        & \hc{0.51}{0.06}$^{\dagger}$ & \hc{0.49}{0.06} & \hc{0.55}{0.06}$^{\dagger}$ & \hc{0.58}{0.06}$^{\dagger}$ & \hc{0.62}{0.06}$^{\dagger}$ & $+0.04$ \\
        \textbf{PyriteRep}& \hc{0.63}{0.06}$^{\dagger}$ & \hc{0.62}{0.06}$^{\dagger}$ & \hc{0.60}{0.06}$^{\dagger}$ & \hc{0.65}{0.06}$^{\dagger}$ & \hc{0.67}{0.06}$^{\dagger}$ & $+0.02$ \\
        \textbf{PyriteRep} w/o Raw 3D         & \hc{0.65}{0.06}$^{\dagger}$ & \hc{0.61}{0.06}$^{\dagger}$ & \hc{0.57}{0.06}$^{\dagger}$ & \hc{0.62}{0.06}$^{\dagger}$ & \textbf{\hc{0.67}{0.06}}$^{\dagger}$ & $+0.05^{*}$ \\
        In-Context                   & \hc{0.62}{0.06}$^{\dagger}$ & \hc{0.58}{0.06}$^{\dagger}$ & \hc{0.47}{0.06} & \hc{0.59}{0.06}$^{\dagger}$ & -- & -- \\
        \midrule
        \multicolumn{7}{l}{\textit{All} ($n$=484)} \\
        \addlinespace[2pt]
        Raw 3D Coord Only            & \hc{0.50}{0.04} & \hc{0.36}{0.05} & \hc{0.50}{0.04} & \hc{0.52}{0.04} & \hc{0.52}{0.04} & $+0.00$ \\
        Interactions + Raw 3D        & \hc{0.60}{0.04} & \hc{0.55}{0.05} & \hc{0.61}{0.04} & \hc{0.64}{0.04} & \hc{0.64}{0.04} & $+0.00$ \\
        \textbf{PyriteRep} & \hc{0.65}{0.04} & \hc{0.61}{0.04} & \hc{0.62}{0.04} & \hc{0.65}{0.04} & \hc{0.66}{0.04} & $+0.01$ \\
        \textbf{PyriteRep} w/o Raw 3D         & \textbf{\hc{0.67}{0.04}} & \hc{0.62}{0.04} & \hc{0.61}{0.04} & \hc{0.63}{0.04} & \textbf{\hc{0.67}{0.04}} & $+0.04$ \\
        In-Context                   & \hc{0.64}{0.04} & \hc{0.59}{0.04} & \hc{0.48}{0.04} & \hc{0.58}{0.04} & -- & -- \\
        \bottomrule
    \end{tabular}
\end{table}

\section{Experiment Results}
\label{sec:results}
We now present and discuss experiment results for all ablations and comparison studied for both \textbf{PyriteReasoner} and \textbf{PyriteFeaturizer}. Because we use data from the \textit{pre-AF3} split in \textbf{Pyrite} for training (both GRPO and the logistic regression in  \textbf{PyriteFeaturizer}), we evaluate only on the \textit{post-AF3} splits. We report per-pair accuracy for \textit{post-AF3 all} and \textit{post-AF3 curated} separately to enable direct comparison with the curated subset we received expert chemist annotations over.

\subsection{PyriteReasoner: Using PyriteRep improves zero-shot LLM performance}
\label{sec:reason}

Table \ref{tab:llm_zeroshot} shows that when \textbf{PyriteReasoner} is provided only raw 3D coordinates, it performs near or below random chance across all LLMs. This suggests that LLMs struggle to extract geometric signal directly from atomic coordinates. However, supplying explicit interaction information substantially improves performance, and using the entire \textbf{PyriteRep} representation yields further gains, reaching accuracy significantly above random chance. Interestingly, omitting raw 3D coordinates from \textbf{PyriteRep} does not hurt performance. Likewise, including in-context examples offer no consistent benefit, possibly because the retrieved examples may not be sufficiently informative. 

In Figure \ref{fig:human}, we more closely compare the performance of the best frontier LLM,  Opus 4.8,  with human experts. We find that unlike human experts, \textbf{PyriteReasoner} shows a much smaller gap in performance between \textbf{Easy} and \textbf{Hard} pose pairs, and slightly underperforming experts in the \textbf{Easy} regime (see Section \ref{compare} for \textbf{Easy}/\textbf{Hard} classification).  Furthermore, upon examining the reasoning traces, we find that  \textbf{PyriteReasoner} is more effective at accurately reasoning about factors like torsion, hydrogen bonds, and MM-GBSA scores in comparison with experts, as well as more likely to explicitly reference domain knowledge about the target protein and ligand (Figure \ref{fig:human} B). Overall, our results show that  \textbf{PyriteReasoner} is able to achieve stronger pose ranking accuracy than expert chemists by not over-relying on heuristics that are adversarial for ranking the high-quality poses in  \textbf{Pyrite}.

\input{coderesults}

\input{grporesults}

%% file: coderesults.tex
\subsection{Classical machine learning is competitive with text-based reasoning}

We next evaluate classical machine learning algorithms on our benchmark using the quantitative features described in Section~\ref{feature}. We compare logistic regression~\citep{cox1958regression} against three tree-based methods capable of capturing nonlinear interactions between features: random forest~\citep{breiman2001random}, LightGBM~\citep{ke2017lightgbm}, and XGBoost~\citep{chen2016xgboost}. 

As shown in Table~\ref{tab:classicalalg}, we observe no statistically significant differences between logistic regression and the tree-based methods on \textit{post-AF3} pose pairs in \textbf{Pyrite}. Notably, even vanilla logistic regression achieves performance comparable to \textbf{PyriteReasoner}, outperforming both Sonnet 5 and GPT-5.6-sol on the \textit{post-AF3 curated} subset (64\% mean per-pose accuracy), although it remains below the strongest LLM, Opus 4.8. These findings highlight the predictive value of quantitative structural features and demonstrate that relatively simple statistical models can be competitive with LLM-based reasoning. This motivates \textbf{PyriteFeaturizer}, which investigates whether LLM-guided feature discovery can further improve the predictive capabilities of simple models by incorporating chemical knowledge beyond existing quantitative features.

\begin{table}[t]
    \centering
\caption{\textbf{PyriteFeaturizer} pose ranking accuracy across four different base LLMs, give input ablations, and  Qwen3.8-27B finetuned with GRPO. Values are mean accuracy $\pm$ 95\% confidence intervals. $\dagger$ means significantly higher than expert, curated split only, and * means significant difference for $\Delta$ GRPO, $p < 0.05$.}
\vspace{5pt}
    \small
\begin{tabular}{lcccccc}
        \toprule
        Arm & Opus 4.8 & Sonnet 5 & GPT-5.6-sol & Qwen3.8-27B & Qwen3.8-27B-GRPO & $\Delta$ \\
        \midrule
        \multicolumn{7}{l}{\textit{Curated} (n=265), $\dagger$ means significantly higher than expert} \\
        \addlinespace[2pt]
        Code Only              & \hc{0.65}{0.06}$^{\dagger}$ & \hc{0.63}{0.06}$^{\dagger}$ & \hc{0.65}{0.06}$^{\dagger}$ & \hc{0.52}{0.06}$^{\dagger}$ & \hc{0.69}{0.06}$^{\dagger}$ & $+0.17^{*}$ \\
        Protein LLMSubcall     & \hc{0.69}{0.06}$^{\dagger}$ & \hc{0.72}{0.05}$^{\dagger}$ & \hc{0.66}{0.06}$^{\dagger}$ & \hc{0.55}{0.06}$^{\dagger}$ & \hc{0.70}{0.06}$^{\dagger}$ & $+0.15^{*}$ \\
        Pose LLMSubcall     & \hc{0.69}{0.06}$^{\dagger}$ & \hc{0.73}{0.06}$^{\dagger}$ & \hc{0.71}{0.06}$^{\dagger}$ & \hc{0.59}{0.06}$^{\dagger}$ & \hc{0.74}{0.05}$^{\dagger}$ & $+0.15^{*}$ \\
        \addlinespace[2pt]
        \multicolumn{7}{l}{\quad\textit{Vanilla Logistic Regression} (all features in PyriteRep): \hcin{0.64}{0.06}$^{\dagger}$} \\
        \midrule
        \multicolumn{7}{l}{\textit{All} (n=484)} \\
        \addlinespace[2pt]
        Code Only              & \hc{0.66}{0.04} & \hc{0.62}{0.04} & \hc{0.65}{0.04} & \hc{0.47}{0.04} & \hc{0.67}{0.04} & $+0.19^{*}$ \\
        Protein LLMSubcall     & \hc{0.65}{0.04} & \hc{0.69}{0.04} & \hc{0.68}{0.04} & \hc{0.53}{0.04} & \hc{0.66}{0.04} & $+0.13^{*}$ \\
        Pose LLMSubcall     & \hc{0.67}{0.04} & \hc{0.69}{0.04} & \hc{0.68}{0.04} & \hc{0.60}{0.04} & \hc{0.71}{0.04} & $+0.10^{*}$ \\
        \addlinespace[2pt]
        \multicolumn{7}{l}{\quad\textit{Vanilla Logistic Regression} (all features in PyriteRep): \hcin{0.60}{0.04}} \\
        \bottomrule
    \end{tabular}
    \label{tab:codegrpo}
\end{table}

\subsection{PyriteFeaturizer: LLM subcalls improve predictive feature discovery}
\label{sec:code}

As shown in Table~\ref{tab:codegrpo}, \textbf{PyriteFeaturizer} generally outperforms \textbf{PyriteReasoner} across models, with the exception of the open-weight model Qwen3.8-27B, perhaps suggest that advanced frontier model capabilities include the ability to translate chemical reasoning into executable, feature-engineering code. Furthermore, allowing \textbf{PyriteFeaturizer} to incorporate LLM-derived features through subcalls further improves performance in most settings, particularly when subcalls evaluate individual poses (e.g. \textit{``Is the ligand well packed in this pose?''}.

We also find that training logistic regression models on \textbf{PyriteFeaturizer}-generated features generally outperform the Vanilla Logistic Regression baseline, suggesting that LLM-guided feature discovery can improve upon the existing quantitative descriptors in \textbf{PyriteRep}. Together, these results highlight that \textbf{PyriteFeaturizer}  serves as an effective hybrid method by leveraging both the learned relationship of precomputed features from training data and any pretrained chemical reasoning knowledge in the underlying languagee model.   Motivated by the weaker performance of Qwen3.8-27B, we next investigate whether reinforcement learning with Group Relative Policy Optimization (GRPO) can improve the feature-discovery capabilities of open-weight models.

%% file: grporesults.tex
\subsection{GRPO narrows the gap between open-weight and frontier models}
\label{grpo}

\begin{figure}
    \centering
    \includegraphics[width=\linewidth]{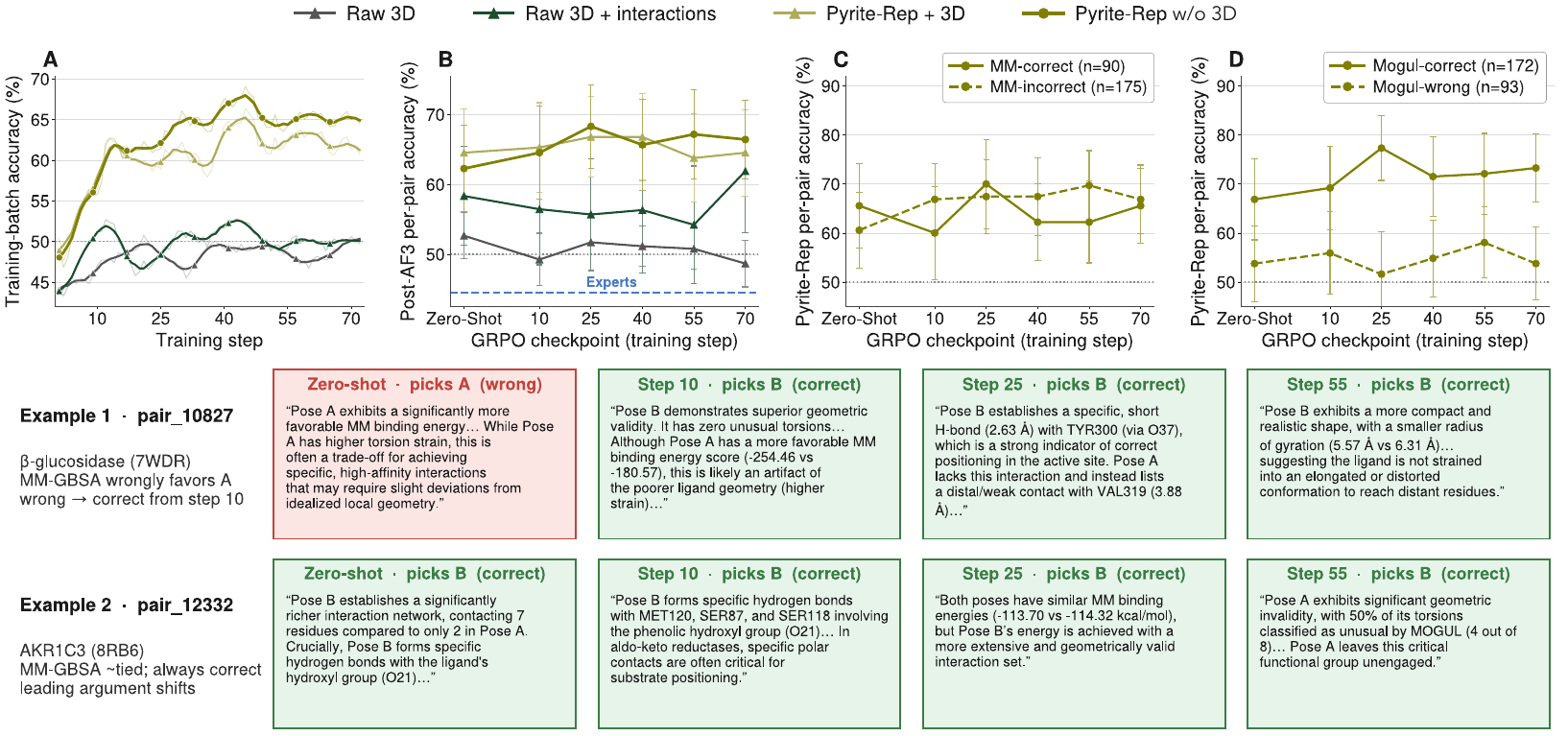}
    \caption{\textbf{GRPO training dynamics for PyriteReasoner}  (A) Train accuracy for four ablations across training steps (error bars are ±SEM) \textbf{(B)} \textit{Post-AF3} per-pair accuracy for four ablations across training steps in comparison with human experts \textbf{(C)} \textit{Post-AF3} per-pair accuracy for only the \textbf{PyriteRep} w/o 3D coordinates ablation, split by   whether MM-GBSA picks incorrect pose. \textbf{(D)} \textit{Post-AF3} per-pair accuracy for only the \textbf{PyriteRep} w/o 3D coordinates ablation, split by   whether Mogul \# Unusual Torsions picks incorrect pose. \textbf{Bottom Rows} Example reasoning  outputs from \textbf{PyriteRep} w/o 3D coordinates ablation over the course GRPO steps.}
    \label{fig:grpo}
\end{figure}

Both \textbf{PyriteReasoner} and \textbf{PyriteFeaturizer} exhibit a performance gap between open-weight and closed-source frontier models, particularly for \textbf{PyriteFeaturizer}. We investigate whether reinforcement learning with Group Relative Policy Optimization (GRPO; Section~\ref{method:grpo}) can improve the pose-ranking capabilities of open-weight models. As shown in Tables~\ref{tab:llm_zeroshot} and~\ref{tab:codegrpo}, GRPO training on \textit{pre-AF3} structures improves performance for both agents, with especially substantial gains for \textbf{PyriteFeaturizer}.

For \textbf{PyriteReasoner}, GRPO improves pose-ranking accuracy on \textit{post-AF3} structures, including the challenging MM-incorrect subset, where the MM-GBSA score favors the incorrect pose. We also observe improvements on cases where the number of unusual Mogul torsions favors the correct pose, suggesting that training improves the model's ability to identify how to rely on different structural features in \textbf{PyriteRep}. Example reasoning traces further illustrate how \textbf{PyriteReasoner} learns to rely less heavily on MM-GBSA over the course of training, suggesting that RL can help refine the chemical heuristics used by LLMs, enabling them to move beyond some of the cognitive biases demonstrated by human experts. 

\begin{figure}
    \centering
\includegraphics[width=\linewidth]{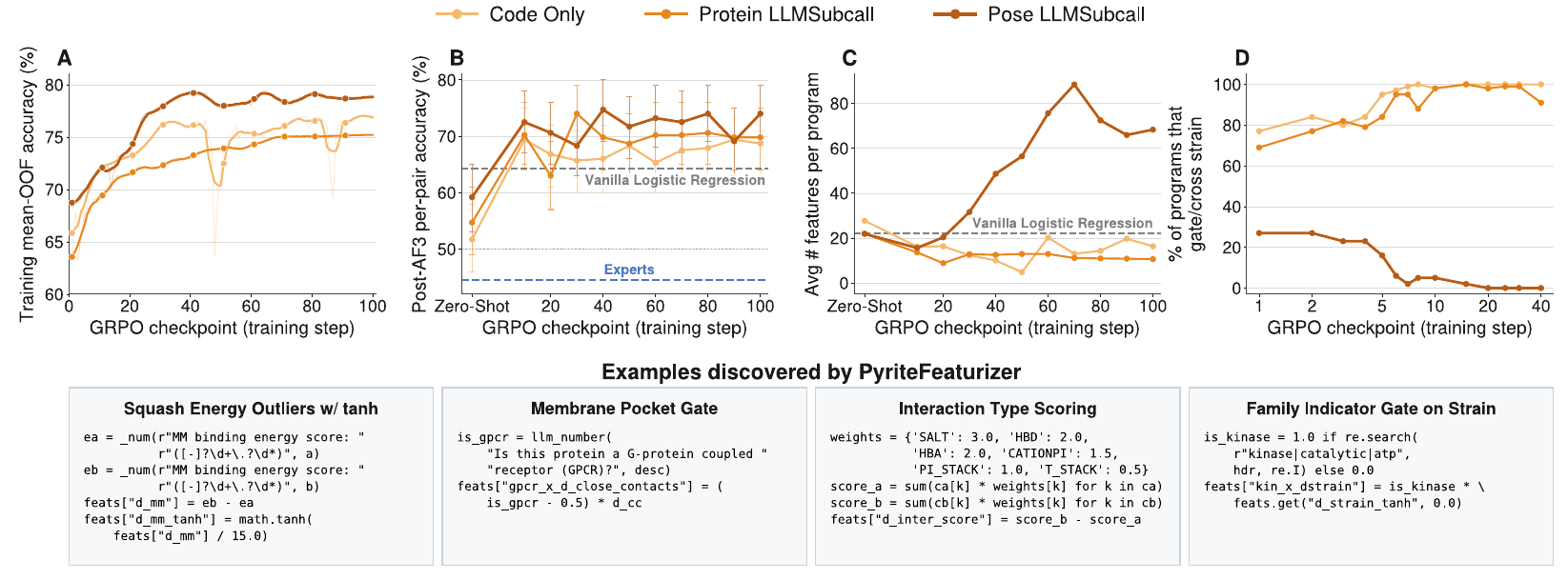}
    \caption{\textbf{GRPO training dynamics for PyriteFeaturizer}  (A) Train mean-OOF accuracy for three ablations across training steps (error bars are ±SEM) \textbf{(B)} \textit{Post-AF3} per-pair accuracy for three ablations across training steps in comparison with human experts and vanilla logistic regression \textbf{(C)} Average \# of distinct features in each program output for all three ablations across training steps \textbf{(D)}  \% of  program outputs using gating logic on strain metrics for all three ablations across training steps. \textbf{Bottom Rows} Example features discovered by  \textbf{PyriteFeaturizer} to use for logistic regression.}
    \label{fig:codegrpo}
\end{figure}

For \textbf{PyriteFeaturizer}, GRPO yields even larger improvements, with the Pose LLM Subcall configuration reaching \textbf{74\%} pose-ranking accuracy, the highest performance observed on our benchmark (Table~\ref{tab:codegrpo}). Figure~\ref{fig:codegrpo} shows increases in both training and \textit{post-AF3} test accuracy across all three configurations, with Pose LLM Subcall achieving the strongest performance. A closer look at the generated programs reveals several learned feature-engineering strategies, including bounding outlier-sensitive quantities such as MM-GBSA scores using \texttt{tanh}, and combining LLM-derived contextual information with quantitative descriptors. For example, one generated feature assigns greater weight to differences in close-contact counts between candidate poses when the target protein is a G protein-coupled receptor (GPCR). We also observe a shift in the feature-generation strategies learned under the Pose LLM Subcall ablation. Rather than relying exclusively on intricate code-based logic, such as conditional gating of ligand strain metrics, the agent increasingly incorporates explicit LLM subcalls to evaluate pose-specific properties. This shift is accompanied by a substantial increase in the number of generated features compared to the other two ablations. Example programs from all three configurations are provided in Appendix~\ref{sec:feature-ex}.

Overall, our results suggest that LLMs may be more effective at pose ranking when used to discover predictive structural features rather than to make direct text-based judgments over each instance in \textbf{Pyrite}. By combining LLM-derived chemical knowledge with a simple classical model, \textbf{PyriteFeaturizer} provides a promising alternative to conventional LLM reasoning. Furthermore, the substantial gains from GRPO demonstrate that RL can enable open-weight models to discover increasingly effective feature representations, even reversing the performance gap with closed-source frontier models. Together, these findings highlight the potential of combining LLM-guided feature discovery with RL for robust and interpretable pose ranking.

%% file: related.tex
\section{Related Work}

\paragraph{Docking and cofolding methods}
Classical docking combines physics-based search with scoring functions to predict binding poses \citep{Trott2009,Koes2013,Friesner2004}. The ability to select near-native poses from fixed candidate sets is evaluated explicitly by the docking-power benchmark in CASF-2016 \citep{su2019casf}. Deep learning has since been applied to both scoring and sampling, though these methods largely retain the rigid-protein assumption \citep{McNutt2025,unimol,corso2023diffdock}. Cofolding models remove this constraint, predicting protein and ligand structure jointly without requiring an experimental structure or known binding mode \citep{Abramson2024,Wohlwend2024}. Like earlier generative models for docking and folding \citep{corso2023diffdock,Jumper2021,Krishna2024}, cofolding models include confidence heads trained to predict error in their own outputs, but recent benchmarks highlight limitations of these confidence scores for ranking diverse ligand poses \citep{pearl,Skrinjar2026-oy}.

\paragraph{LLMs for scientific reasoning}
Large language models are widely used to rank and critique outputs in domains such as writing, mathematics, and coding \citep{llm-judge,math-ranker,coding-ranker}, and agentic systems for scientific discovery rely on these abilities to generate, evaluate, and refine hypotheses \citep{Ren2025-pi}. Several recent benchmarks have shown a steadily improving performance of state-of-the-art LLMs on scientific tasks, including drug discovery \citep{smddbench,Narayanan2025-fm,Mirza2024-os}, and tool use can substantially boost performance even on basic chemistry tasks such as interpreting SMILES \citep{Jang2025-lm,Yu2025-yd}. However, few evaluations target tasks that require reasoning over 3D molecular geometry. Pose ranking offers a well-defined, verifiable instance of this problem.

\paragraph{LLM-based feature engineering and program search}
Several works have leveraged LLMs to support predictive pipelines by generating features for downstream models. For example, CAAFE \citep{hollmann2023caafe} creates contextual feature transformations with executable Python code, while FeatLLM \citep{han2024featllm} combines LLM-generated features with lightweight predictors. FeRG-LLM \citep{ko2025ferg} further trains feature-generating models using supervised learning and preference optimization driven by downstream predictive performance. More broadly, FunSearch \citep{romeraparedes2024funsearch} searches over LLM-generated programs using an automated evaluator, while Eureka \citep{ma2024eureka}  refines reward-function code using feedback from policy training . \textbf{PyriteFeaturizer} applies these ideas to pose ranking, using GRPO training with cross-validation accuracy as a reward to improving LLM-derived features.
\paragraph{Expert judgment in molecular discovery} Previous studies of molecular compound prioritization have similarly demonstrated human bias among medicinal chemists, including differences between reported and statistically inferred decision criteria \citep{kutchukian2012bias}. More broader reviews of virtual screening have documented methodological pitfalls \citep{scior2012pitfalls} that motivate our empirical evaluation of expert's chemical judgment and use of computational tools in the context of pose ranking. 

%% file: conclusion.tex
\section{Future Work \& Conclusion}

AI agents for scientific discovery  ultimately need do more than call the right tool: they
must know how to reason over and when to trust different tools' output. Our pose ranking benchmark \textbf{Pyrite} makes this tension concrete: expert
chemists perform near or below chance, while
physics-based scores can be actively anti-correlated with correctness on our strong,
diverse pose decoys.  Although we showed that both \textbf{PyriteReasoner} and \textbf{PyriteFeaturizer} agents can not only successfully learn from different outputs of computational chemistry tools, but also improve their performance with reinforcement learning, there is still a significant accuracy gap remaining before reaching benchmark saturation on \textbf{Pyrite}. One direction for future work is creating larger training datasets;  we limited our training data to \textit{pre-AF3} structures to match the broader literature, but there might be inherent distribution shifts between \textit{pre-AF3} and \textit{post-AF3} structures that hinder learning methods. Future work should also test whether our
findings hold for poses sampled from other generative models with different sampling techniques, as well as whether improvement on our pairwise pose comparison task reliably translates to pose ranking pipelines used in drug discovery.

%% file: appendix.tex
\section{Pose Pair Filtering Conditions for Pyrite}
\label{sec:filter}
\begin{tcolorbox}[
    colback=purple!10,
    colframe=purple!50,
    title=Pose Pair Filtering Conditions,
    colbacktitle=purple!90,
    coltitle=white
]

\begin{enumerate}[leftmargin=5pt]
    \item \textbf{Ground-truth ligand poorly resolved:} The deposited crystal
    ligand has density support $< 0.75$ RSCC given the observed density map, so the ground truth itself is unreliable.
    \item \textbf{GOOD pose unsupported by density:} The GOOD pose's density support
    is $< 0.50$.
    \item \textbf{Density cannot discriminate the poses:} The two poses'
    density support differ by $< 0.10$.
    \item \textbf{Poses too similar in 3D:} The combined shape-and-chemistry overlap
    score between the two poses $> 1.5$, using OpenEYE ROCS \citep{hawkins2007comparison}.
    \item \textbf{Missing ion from cofolding:} A metal or halide ion that coordinate
    the ligand in the crystal (within $6\,\text{\AA}$) is absent from the cofolded
    structure, so the pose cannot be scored fairly.
    \item \textbf{Protein misaligned:} The C$\alpha$ RMSD between the two poses' protein
    conformations $> 10\,\text{\AA}$.
    \item \textbf{Ligand poses too similar:} The RMSD between the two ligand poses
    $< 1.0\,\text{\AA}$.
    \item \textbf{Too few atoms displaced between poses:} Fewer than $6$ ligand atoms (per ligand
    copy) move by $\geq 2\,\text{\AA}$ between the two poses.
    \item \textbf{Not drug-like:} The ligand fails at least one drug-likeness criterion: QED $\geq 0.29$ \citep{bickerton2012qed}, molecular weight between $100$ and $1000$\,Da, at most $12$ rotatable bonds, and exclusion of simple alkanes.

\end{enumerate}
\end{tcolorbox}

\section{Classical Methods}
\label{sec:classical}

We compare logistic regression against three tree-based methods: Random Forest, XGBoost, and LightGBM. All models are trained on the quantitative features in \textbf{PyriteRep}, using only structures from the \textit{pre-AF3} split, and evaluated on the \textit{post-AF3} splits. Unlike logistic regression, tree-based methods can capture nonlinear relationships and interactions between features without explicit feature engineering. As shown in Table~\ref{tab:classicalalg}, all four methods achieve comparable pose-ranking accuracy, with no statistically significant differences. This suggests that more complex nonlinear models provide limited additional benefit over logistic regression when using the existing quantitative features in \textbf{PyriteRep}.

\begin{table}[h]
    \centering
    \small
    \begin{tabular}{lcc}
        \toprule
        Method & post-AF3 curated & post-AF3 full \\
        \midrule
        Logistic Regression & \hc{0.64}{0.06} & \hc{0.60}{0.04} \\
        Random Forest       & \hc{0.61}{0.06} & \hc{0.65}{0.04} \\
        XGBoost             & \hc{0.59}{0.06} & \hc{0.62}{0.04} \\
        LightGBM            & \hc{0.62}{0.06} & \hc{0.62}{0.04} \\
        \bottomrule
    \end{tabular}
    \caption{Pose-ranking accuracy of classical machine learning methods trained on quantitative features from \textbf{PyriteRep}. Values are mean accuracy $\pm$ 95\% confidence intervals on the \textit{post-AF3 curated} and \textit{post-AF3 full} evaluation splits.}
    \label{tab:classicalalg}
\end{table}

\section{Example Prompt for PyriteReasoner}
\label{prompt-reasoner}
\begin{Verbatim}[breaklines=true, breakanywhere=true,commandchars=\\\{\}]
    ========== SYSTEM PROMPT ==========
You are an expert structural chemist. You will be presented with a pair of protein-ligand poses. Many of these poses have been predicted by different pose-generation models. For each pair, your goal is to choose which pose is closer to the crystallographic binding mode. Please return your SELECTION: A choice from 1-5 corresponding to which pose you believe is closer to the crystallographic binding mode (1= Pose A clearly better, 2= Pose A slightly better, 3= About the same, 4= Pose B slightly better, 5=Pose B clearly better).
POINTS TO KEEP IN MIND: 1. There may exist water-mediated interactions with the structures shown. We do not show them, but expect you to infer them when possible. 2. Please use Option 3= About the same sparingly. 3. We provide the UniProt group name and description for each pose, which you are encouraged in your reasoning and selection. 4. The protein and ligand are the exact same for both poses. 5. You are encouraged to use the provided information -- the raw 3D coordinates, and, when shown, the listed protein-ligand interactions, energetic and strain/validity metrics (e.g. MM binding energy, torsion strain, mogul torsion/ring validity), pose-shape descriptors (radius of gyration, asphericity, extent, internal close contacts, and residue/sidechain contact breadth), and a precomputed structural-features block -- to compare the two poses.

========== USER MESSAGE ==========
UniProt: Aclacinomycin methylesterase RdmC
Description: Involved in the biosynthesis of the anthracycline aclacinomycin which is an aromatic polyketide antibiotic that exhibits high cytotoxicity and is widely applied in the chemotherapy of a variety of cancers. Catalyzes the removal of the methoxy group from the C-15 position of aclacinomycin T and A to yield 15-demethoxyaclacinomycin T and A, respectively.

=== POSE A ===
LIGAND POSE (compact; each section names its format, then lists rows).
atoms [37] (id = element symbol + index): O0 C1 C2 C3 C4 C5 C6 C7 O8 C9 O10 C11 C12 C13 O14 C15 C16 C17 C18 C19 C20 C21 O22 C23 C24 O25 C26 C27 C28 N29 C30 C31 C32 O33 C34 O35 C36
bonds [41] (A-B single, A=B double, A\#B triple, A:B aromatic): C7-O8 C9=O10 C5:C7 C6:C7 N29-C31 C4:C5 C13-O14 N29-C30 C6-C9 C9-C12 C2:C6 C3:C4 C28-N29 C12:C13 C13:C18 C11:C12 C27-C28 C26-C27 C2:C3 C1-C2 C28-C32 C18-C19 C19-C20 C19-O25 C16:C18 C1-C11 C11:C15 O0=C1 O25-C26 C26-O35 C20-C21 C15:C16 C16-C17 C32-O33 C32-C34 C34-O35 C17-C21 C21-C23 C21-O22 C34-C36 C23-C24
coords (id x,y,z, Angstrom): O0 8.532,16.096,11.072  C1 9.339,15.614,10.341  C2 10.104,16.44,9.565  C3 10.033,17.791,9.701  C4 10.811,18.651,8.969  C5 11.675,18.135,8.108  C6 10.989,15.935,8.679  C7 11.764,16.801,7.908  O8 12.583,16.342,7.057  C9 11.101,14.55,8.554  O10 11.917,14.073,7.74  C11 9.446,14.228,10.266  C12 10.333,13.708,9.372  C13 10.463,12.342,9.295  O14 11.296,11.831,8.459  C15 8.702,13.404,11.065  C16 8.83,12.056,10.999  C17 7.989,11.222,11.884  C18 9.709,11.528,10.133  C19 10.013,10.122,10.057  C20 8.968,9.229,10.695  C21 8.478,9.812,12.027  O22 9.527,9.829,12.93  C23 7.392,8.929,12.584  C24 6.84,9.437,13.864  O25 11.218,10.099,10.742  C26 12.252,9.61,10.388  C27 13.264,10.032,9.403  C28 13.958,11.228,9.911  N29 15.083,11.622,8.991  C30 15.985,10.668,8.55  C31 14.614,12.5,7.95  C32 14.456,10.97,11.307  O33 15.499,10.092,11.248  C34 13.396,10.527,12.198  O35 12.914,9.436,11.663  C36 13.845,10.184,13.532
formal\_charges (id:charge, nonzero only): N29:1
target\_smiles: CC[C@]1(O)Cc2cc3c(c(O)c2[C@@H](O[C@H]2C[C@H](N(C)C)[C@H](O)[C@H](C)O2)C1)C(=O)c1c(O)cccc1C3=O
target\_iupac: (7S,9S)-7-[(2R,4S,5S,6S)-4-(dimethylamino)-5-hydroxy-6-methyloxan-2-yl]oxy-9-ethyl-4,6,9-trihydroxy-8,10-dihydro-7H-tetracene-5,12-dione
interactions (RESIDUE-BB|SC-TYPE; e.g. HBD/HBA/ionic/pi): ASP135-SALT ASP135-SC-HBA PHE134-PI_STACK THR106-SC-HBD
MM binding energy score: -241.96
Strain, total (kcal/mol): backbone_torsion=360.7, sidechain_torsion=180.7, total_torsion=541.4
Strain, avg per residue (kcal/mol/residue): backbone_torsion=1.2, sidechain_torsion=0.8, total_torsion=2.0
Pose quality/validity metrics: pb_valid=1, mogul_num_unusual_torsions=2, mogul_percent_unusual_torsions=18.1818, mogul_total_torsions=11, mogul_torsion_penalty=0.960737, mogul_num_unusual_rings=0
Pose shape descriptors (computed from the 3D coordinates + contacts -- use these to compare compactness, elongation, and contact breadth between the poses): radius_of_gyration=4.04, asphericity=5.89, max_extent=11.50, internal_close_contacts=121.00, residues_contacted=3.00, sidechain_residues=2.00

=== POSE B ===
LIGAND POSE (compact; each section names its format, then lists rows).
atoms [37] (id = element symbol + index): O0 C1 C2 C3 C4 C5 C6 C7 O8 C9 O10 C11 C12 C13 O14 C15 C16 C17 C18 C19 C20 C21 O22 C23 C24 O25 C26 C27 C28 N29 C30 C31 C32 O33 C34 O35 C36
bonds [41] (A-B single, A=B double, A#B triple, A:B aromatic): C7-O8 C9=O10 N29-C31 C13-O14 C5:C7 C6:C7 N29-C30 C6-C9 C9-C12 C4:C5 C28-N29 C27-C28 C26-C27 C2:C6 C12:C13 C13:C18 C11:C12 C18-C19 C19-C20 C19-O25 C28-C32 C3:C4 O25-C26 C26-O35 C16:C18 C20-C21 C2:C3 C1-C2 C1-C11 C11:C15 O0=C1 C32-O33 C15:C16 C16-C17 C34-O35 C32-C34 C21-C23 C17-C21 C21-O22 C23-C24 C34-C36
coords (id x,y,z, Angstrom): O0 9.594,18.224,11.328  C1 10.265,17.744,10.426  C2 11.166,18.579,9.731  C3 11.308,19.904,10.079  C4 12.208,20.726,9.434  C5 12.942,20.254,8.4  C6 11.95,18.072,8.707  C7 12.83,18.934,8.031  O8 13.597,18.52,7.052  C9 11.855,16.698,8.376  O10 12.585,16.262,7.462  C11 10.19,16.367,10.124  C12 10.989,15.835,9.111  C13 10.936,14.481,8.847  O14 11.703,14.006,7.909  C15 9.363,15.536,10.808  C16 9.321,14.167,10.575  C17 8.367,13.313,11.394  C18 10.075,13.647,9.572  C19 10.092,12.208,9.238  C20 8.813,11.485,9.66  C21 8.438,11.814,11.091  O22 9.401,11.283,11.976  C23 7.11,11.132,11.365  C24 6.726,11.157,12.801  O25 11.206,11.722,9.913  C26 11.866,10.621,9.438  C27 13.082,10.937,8.58  C28 14.05,11.776,9.354  N29 15.25,12.072,8.558  C30 16.003,10.944,8.052  C31 14.986,13.041,7.51  C32 14.392,11.1,10.671  O33 15.112,9.967,10.46  C34 13.118,10.782,11.449  O35 12.283,9.992,10.648  C36 13.36,10.038,12.711
formal_charges (id:charge, nonzero only): N29:1
target_smiles: CC[C@]1(O)Cc2cc3c(c(O)c2[C@@H](O[C@H]2C[C@H](N(C)C)[C@H](O)[C@H](C)O2)C1)C(=O)c1c(O)cccc1C3=O
target_iupac: (7S,9S)-7-[(2R,4S,5S,6S)-4-(dimethylamino)-5-hydroxy-6-methyloxan-2-yl]oxy-9-ethyl-4,6,9-trihydroxy-8,10-dihydro-7H-tetracene-5,12-dione
interactions (RESIDUE-BB|SC-TYPE; e.g. HBD/HBA/ionic/pi): ASP135-SALT ASP135-SC-HBA PHE134-PI_STACK THR106-SC-HBD
MM binding energy score: -253.98
Strain, total (kcal/mol): backbone_torsion=361.2, sidechain_torsion=164.5, total_torsion=525.7
Strain, avg per residue (kcal/mol/residue): backbone_torsion=1.2, sidechain_torsion=0.7, total_torsion=1.9
Pose quality/validity metrics: pb_valid=1, mogul_num_unusual_torsions=2, mogul_percent_unusual_torsions=18.1818, mogul_total_torsions=11, mogul_torsion_penalty=0.84604, mogul_num_unusual_rings=0
Pose shape descriptors (computed from the 3D coordinates + contacts -- use these to compare compactness, elongation, and contact breadth between the poses): radius_of_gyration=4.26, asphericity=7.94, max_extent=11.86, internal_close_contacts=116.00, residues_contacted=3.00, sidechain_residues=2.00

=== PRECOMPUTED STRUCTURAL FEATURES (from coordinates + protein structure) ===
Ligand: formula C28H33NO8, 5 rings [6, 6, 6, 6, 6], HBA 9, HBD 4, rotatable bonds 4
Ligand functional groups: hydroxyl×4, phenol×2, ketone×2

[Pose A] functional groups: hydroxyl(O8); hydroxyl(O14); hydroxyl(O22); carboxylic acid(C26,O25,O35); amine(N29); hydroxyl(O33)
[Pose A] interactions (ligand FG per contact, resolved via protein coords):
  ASP135-SC-HBA \ensuremath{\leftrightarrow} amine(N29) via N29  (2.87\AA, H-bond)
  THR106-SC-HBD \ensuremath{\leftrightarrow} hydroxyl(O8) via O8  (3.09\AA, H-bond)

[Pose B] functional groups: hydroxyl(O8); hydroxyl(O14); hydroxyl(O22); carboxylic acid(C26,O25,O35); amine(N29); hydroxyl(O33)
[Pose B] interactions (ligand FG per contact, resolved via protein coords):
  ASP135-SC-HBA \ensuremath{\leftrightarrow} amine(N29) via N29  (2.84\AA, H-bond)
  THR106-SC-HBD \ensuremath{\leftrightarrow} hydroxyl(O8) via O8  (2.89\AA, H-bond)

Pose A vs B comparisons:
  Interactions shared: ASP135-SC-HBA, THR106-SC-HBD
  Only A: none;  Only B: none
  Inter-pose ligand RMSD: 1.99\AA (37 atoms)
  Per-FG displacement (A\ensuremath{\rightarrow}B): hydroxyl(O8)=2.4\AA, hydroxyl(O14)=2.3\AA, hydroxyl(O22)=1.7\AA, carboxylic acid(C26)=1.5\AA, amine(N29)=0.6\AA, hydroxyl(O33)=0.9\AA
  FG spacing changes (\ensuremath{\Delta}>1\AA): hydroxyl(O8)\ensuremath{\leftrightarrow}hydroxyl(O33) 8.1\ensuremath{\rightarrow}9.3\AA; hydroxyl(O8)\ensuremath{\leftrightarrow}amine(N29) 5.7\ensuremath{\rightarrow}6.8\AA
  Metric summary -- favors A: none; favors B: energy,strain,mogul; interaction count A=2 B=2

Respond with:
* 3-5 bullet points comparing the poses. Consider both scalar metrics (binding energy, strain) AND structural features (bond validity, torsion geometry, interaction quality).
* Then a final line: SELECTION: <one integer 1-5>

Scale: 1= Pose A clearly better, 2= A slightly better, 3= About the same, 4= B slightly better, 5= Pose B clearly better.

Keep your response between 200 and 400 words.

\end{Verbatim}

\section{Precomputed Feature Block for \textbf{PyriteReasoner}}
\label{sec:precompute}
\begin{tcolorbox}[
    colback=violet!10,
    colframe=violet!50,
    fontupper=\small,
    title=Additional Precomputed Features for PyriteReasoner,
    colbacktitle=violet!90,
    coltitle=white
]
\begin{enumerate}[leftmargin=5pt]
    \item \textbf{Functional Group List}: List of functional groups and their defining atoms.
    \item \textbf{Functional Group Displacement List}: List of $\lVert \mathbf{c}^B_g - \mathbf{c}^A_g \rVert$ for each functional group's centroid $\mathbf{c}_g$.
    \item \textbf{Functional Group Distortion List}: List of functional group pairs whose separation changes between Pose A and Pose B by $>1$\ \AA.
    \item \textbf{Resolved Interactions}: List of interactions mapped to a ligand functional group (\AA).
    \item \textbf{Interaction Count}: Number of resolved interactions.
    \item \textbf{Shared Interactions List}: List of interactions shared by Pose A and Pose B.
    \item \textbf{Unique Interactions List}: List of interactions unique to either Pose A and B, respectively.
    \item \textbf{\# of Rings}: Total number of rings in molecular structure.
    \item \textbf{Ring Size List}: List of the number of atoms in each ring.
    \item \textbf{\# of Hydrogen-bond Acceptors}: Total number of hydrogen-bond acceptor sites. 
    \item \textbf{\# of Hydrogen-bond Donors}: Total number of hydrogen-bond donor sites. 
    \item \textbf{Metric Summary}: List of features in favor of Pose A or Pose B. 
\end{enumerate}
\end{tcolorbox}

\section{GRPO Hyperparameters}
\label{grpo-hparam}
We fine-tune Qwen3.8-27B using Group Relative Policy Optimization (GRPO), with only the language decoder trained via LoRA. 
Training uses 8$\times$H200 GPUs with fully-sharded data parallelism (FSDP) as implemented in the Hugging Face \texttt{trl GRPOTrainer} (v1.9.1)~\citep{vonwerra2020trl}, and \texttt{vLLM} (v0.23.0) for on-policy generation with tensor parallelism across all 8 GPUs using \textit{collocate} mode~\citep{kwon2023vllm}. 
Since vLLM and Hugging Face model forward generates different log-probabilities, we apply importance sampling corrections and off-policy sequence masking, and ensured the importance sampling ratio stays relatively stable throughout training.

For PyriteReasoner, the reward is rule-based: a completion receives 0 unless it ends with a parseable \texttt{SELECTION} line, 1 if it is well-formatted but selects the incorrect pose, and 2 if it selects the correct pose.
For PyriteFeaturizer, the policy writes a Python featurizer that maps the pose-pair text to a set of numeric features. The reward is the accuracy of a logistic regression trained on those features, under 5-fold leave-structure-out cross-validation on the training pool. A featurizer that fails to run, exceeds 20\,s, or yields only constant features receives 0.
To discourage overly long completions, we add the DAPO soft overlong penalty~\citep{yu2025dapo}: a penalty that grows linearly from 0 to $-1$ over the final 3{,}072 tokens before the 10{,}240-token cap, weighted $3\times$.
PyriteFeaturizer is trained for 100 steps of 8 prompts $\times$ 16 completions, without a KL penalty or vLLM importance-sampling correction (Table~\ref{tab:grpo-hyperparams}).

\begin{table}[h]
    \centering
    \caption{GRPO training hyperparameters for PyriteReasoner and PyriteFeaturizer (Qwen3.8-27B).}
    \label{tab:grpo-hyperparams}
    \begin{tabular}{lll}
    \toprule
    Parameter & PyriteReasoner & PyriteFeaturizer \\
    \midrule
    \multicolumn{3}{l}{\textit{Model}} \\
    Base model & Qwen3.8-27B (decoder-only) & Qwen3.8-27B (decoder-only) \\
    LoRA rank / alpha & 16 / 32 & 16 / 32 \\
    LoRA dropout & 0.0 & 0.0 \\
    LoRA targets & q/k/v/o, gate/up/down proj & q/k/v/o, gate/up/down proj \\
    Trainable parameters & 79.7M / 27B (0.29\%) & 79.7M / 27B (0.29\%) \\
    \midrule
    \multicolumn{3}{l}{\textit{Optimization}} \\
    Learning rate & $1 \times 10^{-4}$, cosine, 10\% warmup & $1 \times 10^{-4}$, cosine, 3\% warmup \\
    Prompts per step & 32 & 8 \\
    Global batch size (completions) & 256 & 128 \\
    Training steps & 72 (2 epochs) & 100 \\
    \midrule
    \multicolumn{3}{l}{\textit{GRPO}} \\
    Loss type & DAPO~\citep{yu2025dapo} & DAPO~\citep{yu2025dapo} \\
    KL coefficient $\beta$ & 0.04 & 0 \\
    Clip $\epsilon$ & 0.2 & 0.2 \\
    Group size & 8 & 16 \\
    Updates per batch & 1 (on-policy) & 1 (on-policy) \\
    Entropy bonus & 0.1, adaptive (target 0.5) & None \\
    Sampling temperature & 1.0 & 1.0 \\
    Max completion length & 2048 tokens & 10{,}240 tokens \\
    Mask truncated completions & Yes & Yes \\
    vLLM IS correction & Token-level, truncated at 2.0 & None \\
    Off-policy sequence masking & $\delta = 0.5$ & $\delta = 0.5$ \\
    Thinking mode & Off & Off \\
    Subcall model & -- & Qwen3.5-9B \\
    \midrule
    \multicolumn{3}{l}{\textit{Infrastructure}} \\
    Hardware & 8$\times$H200 & 8$\times$H200 \\
    TRL parallelism & 8-GPU FSDP (full shard) & 8-GPU FSDP (full shard) \\
    vLLM parallelism & 8-GPU TP, colocate & 8-GPU TP, colocate \\
    vLLM max model length & 8192 tokens (14{,}336 for 3D formats) & 14{,}336 tokens \\
    \bottomrule
    \end{tabular}
\end{table}

\section{PyriteFeaturizer Examples}
\label{sec:feature-ex}

\subsection{Code Only}
\begin{Verbatim}[breaklines=true, breakanywhere=true,commandchars=\\\{\}]
import re, math

def featurize(pair_text):
    feats = {}
    try:
        a, _, rest = pair_text.partition("=== POSE B ===")
        b = rest
        ua = a.upper(); ub = b.upper()

        def sd(key, pp, scale=1.0, invert=True):
            # larger = evidence for A
            va = _num(pp, a)
            vb = _num(pp, b)
            d = vb - va  # lower A is better when invert
            if invert is False:
                d = va - vb
            return d / scale

        # --- core strain (lower better) ---
        try: feats["d_strain"] = sd(r"total_torsion=([-]?\textbackslash{}d+\textbackslash{}.?\textbackslash{}d*)", r"", 1.0, True)
        except Exception: pass
        try: feats["d_strain_total2"] = _num(r"total_torsion=([-]?\textbackslash{}d+\textbackslash{}.?\textbackslash{}d*)", b) - _num(r"total_torsion=([-]?\textbackslash{}d+\textbackslash{}.?\textbackslash{}d*)", a)
        except Exception: pass
        try: feats["d_strain_tanh"] = math.tanh((_num(r"total_torsion=([-]?\textbackslash{}d+\textbackslash{}.?\textbackslash{}d*)", b) - _num(r"total_torsion=([-]?\textbackslash{}d+\textbackslash{}.?\textbackslash{}d*)", a)) / 5.0)
        except Exception: pass

        # --- MM energy (more negative better for A) ---
        try:
            ea = _num(r"MM binding energy score: ([-]?\textbackslash{}d+\textbackslash{}.?\textbackslash{}d*)", a)
            eb = _num(r"MM binding energy score: ([-]?\textbackslash{}d+\textbackslash{}.?\textbackslash{}d*)", b)
            feats["d_mm"] = eb - ea  # A more negative -> eb-ea >0 -> favors A
            feats["d_mm_tanh"] = math.tanh(feats["d_mm"] / 15.0)
        except Exception: pass

        # --- MOGUL validity (lower unusual = better A) ---
        try: feats["d_unusual_tor"] = _num(r"mogul_num_unusual_torsions=(\textbackslash{}d+)", b) - _num(r"mogul_num_unusual_torsions=(\textbackslash{}d+)", a)
        except Exception: pass
        try: feats["d_pct_unusual"] = _num(r"mogul_percent_unusual_torsions=(\textbackslash{}d+\textbackslash{}.?\textbackslash{}d*)", b) - _num(r"mogul_percent_unusual_torsions=(\textbackslash{}d+\textbackslash{}.?\textbackslash{}d*)", a)
        except Exception: pass
        try: feats["d_tor_penalty"] = _num(r"mogul_torsion_penalty=(\textbackslash{}d+\textbackslash{}.?\textbackslash{}d*)", b) - _num(r"mogul_torsion_penalty=(\textbackslash{}d+\textbackslash{}.?\textbackslash{}d*)", a)
        except Exception: pass
        try: feats["d_unusual_rings"] = _num(r"mogul_num_unusual_rings=(\textbackslash{}d+)", b) - _num(r"mogul_num_unusual_rings=(\textbackslash{}d+)", a)
        except Exception: pass

        # --- interaction quality: count weighted by type in precomputed blocks ---
        def _iq_text(blk):
            out = {}
            for t, w in [("H-bond", 3.0), ("long/weak", 1.0), ("distal", 0.3)]:
                out[t] = len(re.findall(re.escape(t), blk))
            # near counts by residue interaction tokens
            return out
        try:
            ia = _iq_text(a); ib = _iq_text(b)
            qa = sum(ia[k] * w for k, w in [("H-bond",3.0),("long/weak",1.0),("distal -- verify",0.3)])
            qb = sum(ib[k] * w for k, w in [("H-bond",3.0),("long/weak",1.0),("distal -- verify",0.3)])
            feats["d_iquality"] = qb - qa
        except Exception: pass

        try:
            ca = len(re.findall(r"RES\textbackslash{}d", a)); cb = len(re.findall(r"RES\textbackslash{}d", b))
            feats["d_incount"] = cb - ca
            feats["d_incount_tanh"] = math.tanh((cb - ca) / 2.0)
        except Exception: pass

        # --- intensive: strain per rotatable bond ---
        try:
            rb = _num(r"rotatable bonds (\textbackslash{}d+)", pair_text) or _num(r"rotatable bonds\textbackslash{}s*(\textbackslash{}d+)", pair_text)
            if rb and rb > 0:
                feats["strain_per_rb"] = feats.get("d_strain_total2", 0.0) / rb
        except Exception: pass

        # --- CROSS: kinase indicator * strain delta (strain matters more for kinases) ---
        try:
            hdr = pair_text.split("=== POSE")[0]
            is_kinase = 1.0 if re.search(r"kinase|catalytic|atp", hdr, re.I) else 0.0
            feats["kin_x_dstrain"] = is_kinase * feats.get("d_strain_tanh", 0.0)
            feats["kin_x_dmm"] = is_kinase * feats.get("d_mm_tanh", 0.0)
        except Exception: pass

        # --- CROSS: salt/cationpi unique contacts favor their pose ---
        try:
            salt_a = len(re.findall(r"SALT|CATIONPI", a)); salt_b = len(re.findall(r"SALT|CATIONPI", b))
            feats["d_salt"] = salt_b - salt_a
        except Exception: pass

        # --- signed non-finite / inf guard ---
        try:
            inf_a = 1.0 if ("inf" in a.lower()) else 0.0
            inf_b = 1.0 if ("inf" in b.lower()) else 0.0
            feats["inf_gate"] = inf_b - inf_a  # B surprising -> favors A
        except Exception: pass

    except Exception:
        pass
    # ensure all finite floats
    out = {}
    for k, v in feats.items():
        try:
            v = float(v)
            if math.isfinite(v):
                out[k] = v
            else:
                out[k] = 0.0
        except Exception:
            pass
    return out

\end{Verbatim}

\subsection{w/ Protein LLM Subcall}

\begin{Verbatim}[breaklines=true, breakanywhere=true,commandchars=\\\{\}]

import re
import math

def featurize(pair_text: str) -> dict[str, float]:
    """Parse ONE protein-ligand pose-pair description and return a flat dict of
    SIGNED numeric features describing Pose A relative to Pose B.

    Orientation convention: a LARGER value is stronger evidence FOR Pose A
    (i.e. that Pose A is the crystallographic / correct pose).
    """

    feats: dict[str, float] = {}

    # ---- split per-pose text (A vs B) -------------------------------------
    a_text, _, b_text = pair_text.partition("=== POSE B ===")

    # ---- protein-LEVEL LLM priors (CONSTANT per protein; compute ONCE) -----
    # Pull the header / Description text once; reuse for every question.
    _desc = pair_text.split("=== POSE A ===")[0]

    _kinase = 0.5
    _metal = 0.5
    _gpcr = 0.5
    try:
        _kinase = llm_number(
            "Is this protein a kinase or an enzyme with an active catalytic "
            "site that forms hydrogen bonds and salt bridges in a binding "
            "pocket? Answer 1 if yes, 0 if no.",
            _desc,
        )
    except Exception:
        pass
    try:
        _metal = llm_number(
            "Does this protein coordinate a metal ion or contain charged "
            "catalytic residues (acidic/basic) in an active site? Answer 1 "
            "if yes, 0 if no.",
            _desc,
        )
    except Exception:
        pass
    try:
        _gpcr = llm_number(
            "Is this protein a GPCR, ion channel, or other membrane receptor "
            "with a deep binding cavity? Answer 1 if yes, 0 if no.",
            _desc,
        )
    except Exception:
        pass

    # ------------------------------------------------------------------------
    # "lower is better" metrics:  (B - A)  =>  A lower (better)  =>  positive
    # ------------------------------------------------------------------------

    # Strain: lower strain is better.  A lower than B  =>  (B - A) > 0  =>  favors A.
    try:
        feats["d_strain_total"] = (
            _num(r"total_torsion=([-\textbackslash{}d.]+)", b_text)
            - _num(r"total_torsion=([-\textbackslash{}d.]+)", a_text)
        )
    except Exception:
        pass

    # Mogul torsion penalty: lower is better.
    try:
        feats["d_mogul_torsion_penalty"] = (
            _num(r"mogul_torsion_penalty=([-\textbackslash{}d.]+)", b_text)
            - _num(r"mogul_torsion_penalty=([-\textbackslash{}d.]+)", a_text)
        )
    except Exception:
        pass

    # Mogul number of unusual torsions: lower is better.
    try:
        feats["d_mogul_num_unusual"] = (
            _num(r"mogul_num_unusual_torsions=(\textbackslash{}d+)", b_text)
            - _num(r"mogul_num_unusual_torsions=(\textbackslash{}d+)", a_text)
        )
    except Exception:
        pass

    # Internal close contacts: fewer is better.
    try:
        feats["d_internal_close_contacts"] = (
            _num(r"internal_close_contacts=([-\textbackslash{}d.]+)", b_text)
            - _num(r"internal_close_contacts=([-\textbackslash{}d.]+)", a_text)
        )
    except Exception:
        pass

    # Asphericity: lower is more compact / native-like.
    try:
        feats["d_asphericity"] = (
            _num(r"asphericity=([-\textbackslash{}d.]+)", b_text)
            - _num(r"asphericity=([-\textbackslash{}d.]+)", a_text)
        )
    except Exception:
        pass

    # Mogul percent unusual torsions: lower is better.
    try:
        feats["d_mogul_percent_unusual"] = (
            _num(r"mogul_percent_unusual_torsions=([-\textbackslash{}d.]+)", b_text)
            - _num(r"mogul_percent_unusual_torsions=([-\textbackslash{}d.]+)", a_text)
        )
    except Exception:
        pass

    # Mogul number of unusual rings: lower (0) is better.
    try:
        feats["d_mogul_unusual_rings"] = (
            _num(r"mogul_num_unusual_rings=(\textbackslash{}d+)", b_text)
            - _num(r"mogul_num_unusual_rings=(\textbackslash{}d+)", a_text)
        )
    except Exception:
        pass

    # ------------------------------------------------------------------------
    # MM binding energy: MORE NEGATIVE = stronger (better).  We want A to be
    # more negative (lower) to favor A  =>  (B - A):  A lower  =>  positive.
    # (Kept; MM is weak on its own but still a signed signal.)
    # ------------------------------------------------------------------------
    try:
        feats["d_mm_binding_energy"] = (
            _num(r"MM binding energy score: ([-\textbackslash{}d.]+)", b_text)
            - _num(r"MM binding energy score: ([-\textbackslash{}d.]+)", a_text)
        )
    except Exception:
        pass

    # Kinase prior weighting the strain difference.
    try:
        feats["kinase_x_d_strain"] = (
            (_kinase - 0.5) * feats.get("d_strain_total", 0.0)
        )
    except Exception:
        pass

    # Metal / charged-active-site prior weighting the mogul penalty difference.
    try:
        feats["metal_x_d_mogul_penalty"] = (
            (_metal - 0.5) * feats.get("d_mogul_torsion_penalty", 0.0)
        )
    except Exception:
        pass

    # GPCR / cavity prior weighting the close-contact difference.
    try:
        feats["gpcr_x_d_close_contacts"] = (
            (_gpcr - 0.5) * feats.get("d_internal_close_contacts", 0.0)
        )
    except Exception:
        pass

    return feats

\end{Verbatim}

\subsection{w/ Pose LLM Subcall}

\begin{Verbatim}[breaklines=true, breakanywhere=true,commandchars=\\\{\}]
import re, math

def featurize(pair_text):
    feats = {}
    try:
        a, sep1, b = pair_text.partition("=== POSE B ===")
        cmp_sec = ""
        m = re.search(r"Pose A vs B comparisons:", pair_text)
        if m:
            cmp_sec = pair_text[m.start():]
        rm = re.search(r"Metric summary.*?favors A:\textbackslash{}s*([^;]*);?\textbackslash{}s*favors B:", cmp_sec)
        def _f(p, pat, d=0.0):
            v = _num(pat, p)
            return v if math.isfinite(v) else d
        # strain: lower is better -> B - A
        feats["B_minus_A_total_tors"] = _f(b, r"total_torsion=([-\textbackslash{}d.]+)") - _f(a, r"total_torsion=([-\textbackslash{}d.]+)")
        feats["B_minus_A_avg_tors"] = _f(b, r"residue\textbackslash{}): backbone_torsion=([-\textbackslash{}d.]+),[\textbackslash{}s\textbackslash{}S]*?total_torsion=([-\textbackslash{}d.]+)") - _f(a, r"residue\textbackslash{}): backbone_torsion=([-\textbackslash{}d.]+),[\textbackslash{}s\textbackslash{}S]*?total_torsion=([-\textbackslash{}d.]+)")
        # energy more negative = stronger A -> A - B
        try:
            feats["A_minus_B_energy"] = _f(a, r"MM binding energy score:\textbackslash{}s*([-\textbackslash{}d.]+)") - _f(b, r"MM binding energy score:\textbackslash{}s*([-\textbackslash{}d.]+)")
        except Exception:
            pass
        # mogul: lower better -> B - A
        for pg, key in [(r"mogul_num_unusual_torsions=(\textbackslash{}d+)", "d_mogul_unusual_tors"),
                        (r"mogul_num_unusual_rings=(\textbackslash{}d+)", "d_mogul_unusual_rings"),
                        (r"mogul_torsion_penalty=([-\textbackslash{}d.]+)", "d_mogul_penalty")]:
            feats[key] = _f(b, pg) - _f(a, pg)
        # interact quality
        def iq(p):
            s = 0.0
            for mm in re.findall(r"(HBD|HBA|SALT|CATIONPI|PI_STACK|T_STACK)", p):
                w = {"SALT": 2.0, "HBD": 1.2, "HBA": 1.0, "CATIONPI": 0.8, "PI_STACK": 0.7, "T_STACK": 0.5}.get(mm, 0.4)
                s += w
            return s
        feats["A_minus_B_iq"] = iq(a) - iq(b)
        feats["A_minus_B_cnt"] = len(re.findall(r"\textbackslash{}d+-BB|[\textbackslash{}w]+-SC|-\textbackslash{}d+-\textbackslash{}w+-\textbackslash{}w+", a)) - 0
        # LLM per-pose, signed
        for q, key in [
            ("Does this pose make a geometrically ideal, well-satisfied H-bond network?", "llm_hbond"),
            ("Are all polar groups of the ligand satisfied in this pose?", "llm_polar_sat"),
            ("Does this pose show any strained or clashing (unphysically close) contacts?", "llm_clash"),
            ("Is the ligand well packed with good steric complementarity in this pose?", "llm_pack"),
        ]:
            feats["A_minus_B_" + key] = (llm_number(q, a) - llm_number(q, b))
            if key == "llm_clash":
                feats["A_minus_B_" + key] = -feats["A_minus_B_" + key]
        # normalizers
        rbs = _num(r"rotatable bonds\textbackslash{}s+(\textbackslash{}d+)", pair_text, 1.0) or 1.0
        feats["d_tors_per_rb"] = feats.get("B_minus_A_total_tors", 0.0) / rbs
        feats["d_pen_per_rb"] = feats.get("d_mogul_penalty", 0.0) / rbs
    except Exception:
        pass
    for k in list(feats):
        try:
            if not math.isfinite(feats[k]):
                feats[k] = 0.0
        except Exception:
            feats.pop(k, None)
    return feats
    \end{Verbatim}

\section{Comparison with Sign-Flipped MM-GBSA}
\label{sec:app-flip}

Because selecting the pose with the \emph{lower} MM-GBSA score is anti-correlated
with correctness on \textsc{Pyrite} (Table~\ref{tab:classic}), the
complementary rule of selecting the \emph{higher}-scoring pose 
achieves above-chance accuracy. We report its performance in
Table~\ref{tab:flipped-mmgbsa} but do not treat it as a competitive baseline for the following reasons:

\begin{enumerate}
    \item \textbf{The sign is regime-dependent.} MM-GBSA estimates
binding free energy, and the convention that lower values are more favorable
is grounded in the meaning of this quantity, rather than an arbitrary labeling
choice. Figure~\ref{fig:placeholder} shows that the inversion in does not hold
universally: the relationship degrades only in the tight top-$k$ regime from
which \textsc{Pyrite} pairs are drawn. Choosing the inverted sign a priori
therefore requires knowing how aggressively candidates have already been
filtered, which is equivalent to solving the pose-ranking problem itself.
\item \textbf{If the inverted signal were learnable from training data, our methods would have found it.}
Vanilla Logistic Regression and \textsc{PyriteFeaturizer} both receive MM-GBSA as
an input feature with no constraint on coefficient sign, and GRPO rewards ranking
accuracy directly. That these methods instead converge on feature combinations (and,
for \textsc{PyriteFeaturizer}, on transformations bounding MM-GBSA's influence
(Figure~\ref{fig:codegrpo})) indicates  that the inverted heuristic is not the
dominant signal during training. 
\item \textbf{Learning-based methods beat score inversion without imitating it.}
Table~\ref{tab:flipped-mmgbsa} reports accuracy alongside prediction-level
agreement with the inverted heuristic. After GRPO,  \textsc{PyriteReasoner} exceeds
the sign-flipped MM-GBSA score while agreeing with it on only half of pairs. The trained model is therefore not simply reward-hacking \textbf{Pyrite} by converging to an inverted heuristic, but learning a substantially different
and more accurate decision rule. We take this decoupling of accuracy from
agreement as the clearest evidence that sign-inverted MM-GBSA is an
artifact of ranking high-quality poses rather than a reliable and learnable signal.

\end{enumerate}

\begin{table}[h]
\centering
\caption{Accuracy on the post-AF3 full split and prediction-level agreement with
the sign-flipped MM-GBSA rule. \textsc{PyriteReasoner} surpasses the sign-flipped
rule while agreeing with it at close to the chance rate.}
\label{tab:flipped-mmgbsa}
\begin{tabular}{lcc}
\toprule
Method & Accuracy & Agreement with sign-flipped MM-GBSA \\
\midrule
Sign-flipped MM-GBSA            & 64.7\% & 100.0\% \\
\textsc{PyriteReasoner}, zero-shot      & 63.0\% & 46.3\% \\
\textsc{PyriteReasoner}, GRPO step 70   & 67.4\% & 51.4\% \\
\bottomrule
\end{tabular}
\end{table}